\documentclass[a4paper,fleqn]{cas-sc}

\usepackage[authoryear,longnamesfirst]{natbib}

\let\cite\citep  
\usepackage{amsmath,amssymb}
\usepackage{booktabs}
\usepackage{array}
\usepackage{enumitem}
\setlist{nosep,leftmargin=*}

\newcolumntype{L}{l}
\newcolumntype{R}{r}
\newcolumntype{C}{c}

\def\tsc#1{\csdef{#1}{\textsc{\lowercase{#1}}\xspace}}
\tsc{WGM}
\tsc{QE}
\tsc{EP}
\tsc{PMS}
\tsc{BEC}
\tsc{DE}

\usepackage[ruled,linesnumbered,procnumbered]{algorithm2e}
\SetAlFnt{\small}
\SetKw{KwAnd}{and}
\SetKw{KwInWord}{in}
\SetKwFunction{KwNull}{Null}
\SetKwFunction{KwTrue}{True}
\SetKwFunction{KwFalse}{False}

\usepackage{multirow}
\usepackage{makecell}
\usepackage{colortbl}
\usepackage{pifont}

\usepackage{color,xcolor}
\definecolor{Gray}{gray}{0.9}
\definecolor{light-gray}{gray}{0.95}

\usepackage{amsthm}

\theoremstyle{plain}
\newtheorem{lemma}{Lemma}
\newtheorem{theorem}{Theorem}

\newcommand{\ours}{OccLinker\xspace} 

\newcommand{\rev}[1]{#1}  

\begin{document}
\let\WriteBookmarks\relax
\shorttitle{Deflickering Vision-Based Occupancy Networks}
\shortauthors{Fengcheng Yu et~al.}

\title[mode = title]{Deflickering Vision-Based Occupancy Networks through Lightweight Spatio-Temporal Correlation}

\author[1]{Fengcheng Yu}[style=chinese,orcid=0009-0003-0831-9817]
\ead{fyu54107@usc.edu}
\credit{Software, Methodology, Formal analysis, Visualization, Writing – original draft}

\author[1,2]{Haoran Xu}[style=chinese,orcid=0000-0002-9330-2475]
\ead{xuhr9@mail2.sysu.edu.cn}
\credit{Conceptualization, Investigation, Methodology, Visualization, Writing – original draft}

\author[1,2]{Canming Xia}[style=chinese,orcid=0000-0001-7743-1392]
\ead{xiacm@mail2.sysu.edu.cn}
\credit{Writing – review and editing, Validation}

\author[1]{Ziyang Zong}[style=chinese,orcid=0000-0001-6556-1085]
\ead{zongzy@mail2.sysu.edu.cn}
\credit{Writing – review and editing, Validation}

\author[1]{Guang Tan}[style=chinese,orcid=0000-0002-0658-8867]
\ead{tanguang@mail.sysu.edu.cn}
\cormark[1]
\credit{Conceptualization, Writing – review and editing, Supervision, Funding acquisition}

\affiliation[1]{organization={the School of Intelligent Systems Engineering},
                addressline={Shenzhen Campus of Sun Yat-sen University}, 
                city={Shenzhen},
                postcode={518107}, 
                country={China}}
\affiliation[2]{organization={the Department of Intelligent Computing},
                addressline={Peng Cheng Laboratory}, 
                city={Shenzhen},
                postcode={518066}, 
                country={China}}

\cortext[cor1]{Corresponding author}

\nonumnote{
Fengcheng~Yu and Haoran~Xu contributed equally to the paper.
}

\begin{abstract}
Vision-based occupancy networks (VONs) provide an end-to-end solution for reconstructing 3D environments in autonomous driving. However, existing methods often suffer from temporal inconsistencies, manifesting as flickering effects that compromise visual experience and adversely affect decision-making. While recent approaches have incorporated historical data to mitigate the issue, they often incur high computational costs and may introduce noisy information that interferes with object detection. We propose \ours, a novel plugin framework designed to seamlessly integrate with existing VONs for boosting performance. Our method efficiently consolidates historical static and motion cues, learns sparse latent correlations with current features through a dual cross-attention mechanism, and produces correction occupancy components to refine the base network's predictions. We propose a new temporal consistency metric to quantitatively identify flickering effects. Extensive experiments on two benchmark datasets demonstrate that our method delivers superior performance with negligible computational overhead, while effectively eliminating flickering artifacts.
\end{abstract}



\maketitle

\section{Introduction}
\label{sec:intro}

Vision-based Occupancy Networks (VONs)~\cite{OccFormer,VoxFormer} have emerged as a powerful paradigm for reconstructing surrounding environments from ego-centric multi-view images. Typically, VONs represent the scene as a structured grid of voxels~\cite{Occ3D} by learning accurate mappings from 2D visual cues to 3D occupancy representations. This approach enables the extraction of both geometric structure and semantic understanding, thereby facilitating various applications, including indoor navigation~\cite{liu2024volumetric} and autonomous driving~\cite{xu2025exploiting}.

\begin{figure}[pos=!htbp]
\centering
\includegraphics[width=0.5\textwidth]{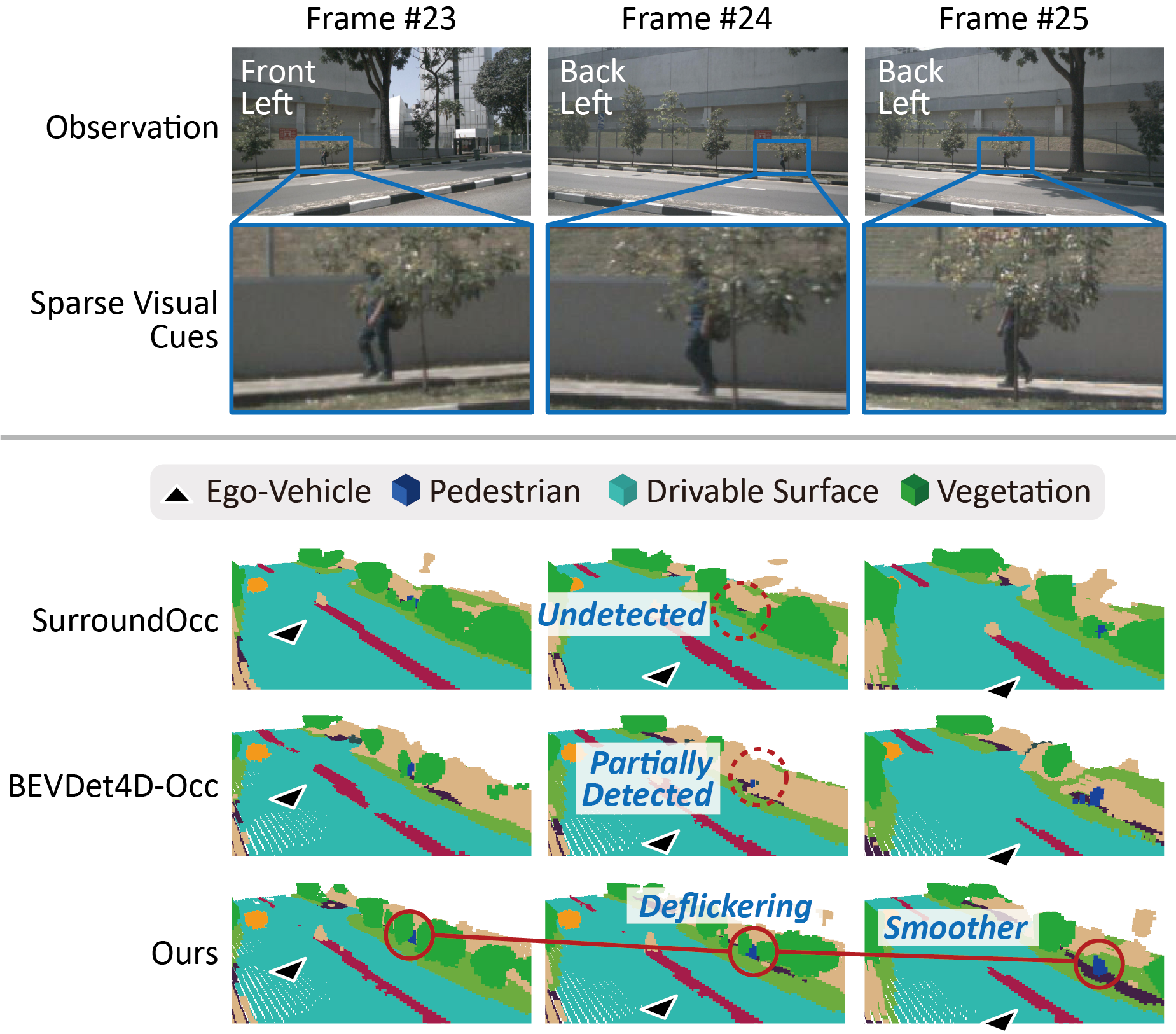}
\caption{Enhancing existing VONs with \ours. (i) In the upper part, the first row shows three consecutive images where the ego-vehicle moves forward along the right side of the road. The blue box highlights a pedestrian as a sparse visual cue, and the second row provides the corresponding zoomed-in views. (ii) In the lower part, we compare different types of VON methods. The first row shows a standard 3D VON~\cite{surroundOcc} that ignores historical information, resulting in the pedestrian being missed in certain frames. The second row presents a history-aware VON~\cite{bevdet4d} that utilizes past frames but still produces incomplete voxels due to suboptimal temporal association. In contrast, our module, when integrated with the 3D VON~\cite{surroundOcc}, achieves smoother and more complete occupancy predictions through effective spatio-temporal correlation.}
\label{fig:moti}
\end{figure}

Despite these progresses, the occupancy results generated by existing VONs often exhibit inaccuracies and visual imperfections. A particularly problematic phenomenon, which we term \textit{\textbf{flickering}}, manifests as instability in the constructed scenes: objects may appear or disappear abruptly, accompanied by various artifacts. This lack of temporal consistency and accuracy not only confuses the driving control system but also seriously degrades the visual experience for human drivers. The primary causes of flickering include sensor noise, occlusion, model limitations, and thresholding issues. As illustrated in \figurename~\ref{fig:moti}, a pedestrian occluded by a tree provides incomplete visual cues, posing a significant challenge to the prediction models. In this case, SurroundOcc~\cite{surroundOcc}, which operates in a frame-by-frame ``detection'' paradigm, fails to detect the pedestrian in Frame \#24. The temporary absence of the person will create a flickering effect when the results are displayed to the user.

\begin{figure}[pos=!htbp]
\centering
\includegraphics[width=0.55\textwidth]{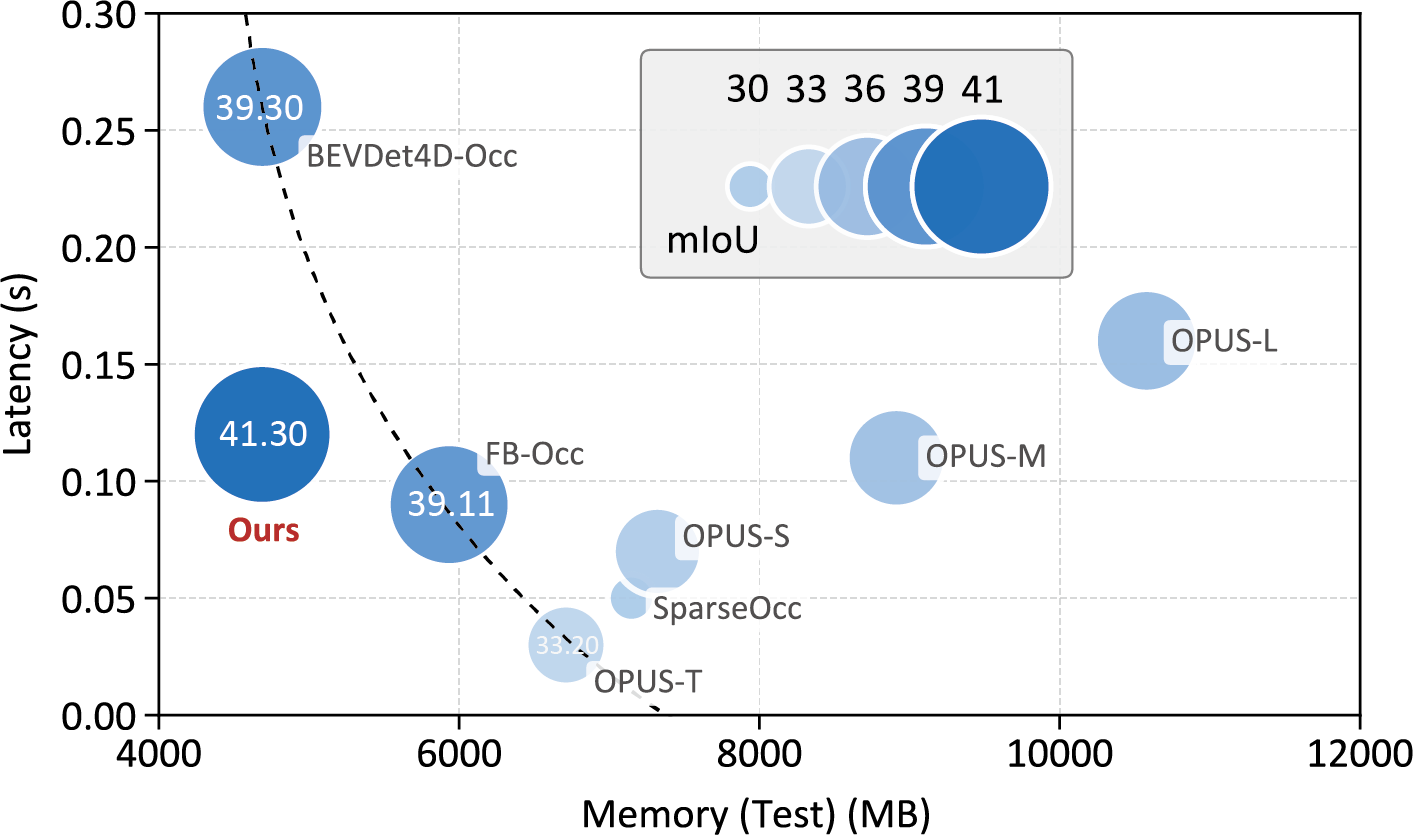}
\caption{Comparison with history-aware VON methods by mIoU and running costs. Points closer to the bottom-left indicate better efficiency, with lower memory usage and faster inference. Larger circles denote higher prediction quality with mIoU shown in numbers. \rev{Our method, integrated with ViewFormer, achieves the best mIoU while keeping memory and latency close to those of the base network.} \rev{Memory and latency in this figure are the L20 measurements of \tablename~\ref{tab:efficiency}.}}
\label{fig:bubble}
\end{figure}

An effective approach to mitigate the temporal inconsistency problem is to incorporate historical information when predicting the current occupancy. Recent history-aware VONs have explored this direction and can be broadly categorized into two classes: dense Bird's Eye View (BEV)/voxel-based temporal fusion methods, exemplified by BEVDet4D-Occ~\cite{bevdet4d} and ViewFormer~\cite{viewformer}, and sparse query or representation based methods, exemplified by OPUS~\cite{opus} and SparseOcc~\cite{SparseOcc_Liu}. Specifically:

\begin{itemize}
    \item \textbf{Dense methods}, such as BEVDet4D-Occ~\cite{bevdet4d}, FB-Occ~\cite{fb_occ} and ViewFormer~\cite{viewformer}, perform dense temporal modeling by propagating, aligning, or aggregating historical BEV features or volumetric representations into the current frame. These approaches incur substantial computational and memory costs, often limiting the feasible temporal window or spatial resolution. For example, as shown in \figurename~\ref{fig:bubble}, BEVDet4D-Occ produces the largest inference latency. Moreover, dense temporal aggregation may propagate redundant or noisy features across broad spatial regions, leading to unstable predictions in dynamic areas. As illustrated in the lower part of \figurename~\ref{fig:moti}, the pedestrian appears unclear in the BEVDet4D-Occ results, likely due to coarse-grained temporal fusion.

    \item \textbf{Sparse methods} seek to reduce the complexity of dense voxel fusion by operating on sparse queries~\cite{opus} or representations~\cite{SparseOcc_Liu,jiang2025gausstr}. By focusing computation on relevant latent regions, methods such as SparseOcc and the OPUS series typically achieve lower inference latency (see \figurename~\ref{fig:bubble}). These methods often require multi-stage recovery or decoding processes to reconstruct dense occupancy. For instance, SparseOcc employs diffusion-based propagation combined with feature pyramids to densify non-empty regions across scales, whereas OPUS adopts a coarse-to-fine cross-attention mechanism to progressively decode sparse queries into 3D space. These designs increase memory usage and reduce voxel-level accuracy.
\end{itemize}

Different from existing methods, we explore an alternative approach: enhancing temporal consistency through a lightweight plug-in module that leverages historical information without redesigning the backbone, altering the fusion strategy, or retraining the base VON from scratch. Specifically, our approach only requires training a lightweight module on top of existing VON predictions or intermediate features, making it both practical and broadly applicable.

To this end, we propose a novel method, termed \textbf{\ours}. Our approach improves occupancy prediction by discriminating and extracting relevant motion and static cues from historical frames, emphasizing \emph{lightweight yet effective} spatio-temporal correlation modeling. \ours follows a three-stage framework. First, it converts historical static and motion cues into queries through sparse tokenization. Next, an independent latent correlation module is introduced to attentively associate current features with historical cues within a shared latent space. Finally, the system generates {\em correction occupancy} that enables refinement of the base network's results. Our method offers two key advantages: 

\begin{itemize}
    \item Seamless integration with state-of-the-art (SOTA) VONs for enhanced prediction accuracy. When integrated with SurroundOcc~\cite{surroundOcc} on its benchmark, it elevates IoU and temporal consistency of moving objects (as defined in Section~\ref{sec:exp}) by 1.63\% and 2.31, respectively. Similarly, when combined with ViewFormer~\cite{viewformer} on the Occ3D~\cite{Occ3D} benchmark, it achieves gains of 0.24\% in IoU and 2.87 in temporal consistency;

    \item Low cost through lightweight correlation operation, which surpasses recent history-aware VONs~\cite{bevdet4d,ouyang2024linkocc} with much lower computational costs. For instance, OPUS-L~\cite{opus} requires 10.33 GB of GPU memory and 0.16 seconds per frame for inference. In contrast, ViewFormer+\ours only needs 4.58 GB memory and 0.12 seconds. Notably, the \ours plug-in itself uses just 105 MB of memory and adds only 0.01 seconds of latency.
\end{itemize}

These two advantages jointly push our method beyond the usual accuracy‑efficiency Pareto frontier. As illustrated in \figurename~\ref{fig:bubble}, existing history‑aware approaches demonstrate a clear trade-off between performance and cost. In contrast, our method, when combined with ViewFormer, achieves superior accuracy without sacrificing efficiency.

In summary, the contributions of this paper are three-fold:
\begin{itemize}
    \item We introduce \ours, a parameter-efficient plug-in that integrates seamlessly with existing 3D VONs without requiring any modifications to the base network. It incurs only minimal GPU memory overhead and negligible inference latency, while consistently enhancing prediction accuracy.

    \item We propose a lightweight spatio-temporal correlation learning method that effectively fuses historical cues, sparse motion features, as well as current static tokens. These cues are mapped into a shared compact latent space, enabling pixel-to-voxel correlation through adaptive 3D occupancy aggregation.
    
    \item We devise a new temporal consistency metric to help analyze the effectiveness of various occupancy methods. Extensive experiments are conducted on both SurroundOcc~\cite{surroundOcc} and Occ3D~\cite{Occ3D} benchmarks, showing the advantages of our approach in accuracy while maintaining real-time inference efficiency.
\end{itemize}


\section{Related Work}
\label{sec:relate}

\subsection{3D Scene Reconstruction}
3D scene reconstruction technology has evolved rapidly in autonomous driving. \rev{Early work recovered geometry from explicit camera models. El Hazzat et al.~\cite{3D_reconstruction_camera_para_3} reconstruct scenes from uncalibrated videos using incremental structure from motion, self-calibration, and bundle adjustment. MonoScene~\cite{MonoScene} infers dense geometry and semantics from a single monocular RGB image and remains a reference for lifting 2D observations into 3D. Learning-based methods further improve reconstruction through implicit geometry and sparse representations. Geo-Neus~\cite{Multi_View_reconstruction_1} constrains the zero-level set of a signed distance function using sparse structure-from-motion geometry and photometric consistency. Closer to our setting, OSP~\cite{shi2024occupancy} replaces dense volumes with points of interest, allowing the model to focus on selected regions and areas beyond the conventional perception range.}


\subsection{3D VONs}
3D VONs extend 3D scene reconstruction by predicting the occupancy and semantics of each voxel from LiDAR, cameras, or both, providing fine-grained environmental perception for autonomous driving. \rev{For LiDAR-based prediction, S3CNet~\cite{S3CNet_depth} uses sparse convolutions for scene completion, while OccMamba~\cite{li2025occmamba} replaces attention with state space models to reduce quadratic complexity. HGS-OCC~\cite{duan2026hgsocc} combines cameras and LiDAR by refining monocular depth with sparse point measurements while maintaining real-time inference.} Camera-only methods such as TPVFormer~\cite{TPVFormer} and VoxFormer~\cite{VoxFormer} typically use Transformers~\cite{attention_is_all_you_need} to learn pixel-to-voxel relationships and extract rich semantic information at a lower sensor cost. However, processing high-dimensional images and dense voxel grids remains computationally expensive. To improve efficiency, recent methods use sparse Gaussians as intermediate representations. \rev{GaussianFormer~\cite{huang2024gaussianformer} achieves accuracy comparable to dense methods using only 17.8\%--24.8\% of their memory, while GaussTR~\cite{jiang2025gausstr} avoids voxel annotations by aligning rendered Gaussian features with foundation models. Other label-efficient methods include OccCLIP~\cite{zhang2025clip}, which supports unseen-class prediction through voxel--text alignment, and PPMNet~\cite{lu2025geometry}, which jointly estimates depth and semantics before lifting them into 3D.}

\subsection{History-aware VONs}
Temporal 3D occupancy prediction generally follows two paradigms: (i) 4D VONs that estimate current and future states, such as Cam4DOcc~\cite{Cam4docc} and OccSora~\cite{OccSora}, and (ii) history-aware VONs that use past observations to refine current predictions. This work focuses on the latter, which includes dense methods such as BEVDet4D-Occ~\cite{bevdet4d}, ViewFormer~\cite{viewformer}, and \rev{TQL-SSC~\cite{zhao2026temporal}}, as well as sparse methods such as SparseOcc~\cite{SparseOcc_Liu} and \rev{LinkOcc~\cite{ouyang2024linkocc}}. BEVDet4D-Occ aligns temporal features in the dense BEV space, while \rev{TQL-SSC exploits historical images to recover regions invisible in the current frame. In the sparse setting, LinkOcc contrastively associates decoder queries across adjacent frames.} Although effective, these methods integrate temporal modeling into their specific architectures and require joint training. In contrast, \ours is a lightweight plugin that uses complementary static and motion cues to refine current predictions without modifying the base architecture. It can be integrated with both single-frame and history-aware VONs to improve occupancy accuracy and temporal consistency.



\section{Methodology}
\label{sec:meth}

\subsection{Problem Formulation}
\label{sec:meth:problem}
In the context of vision-based occupancy networks (VONs)~\cite{xu2025exploiting,VoxFormer,surroundOcc} for autonomous driving, the ego-vehicle typically collects two types of data:
(i) multi-view RGB frames from surround-view cameras (typically six), captured at a relatively low and fixed frequency -- we refer to these as keyframes;
(ii) higher-frequency motion information between keyframes, such as optical flow or frame differences.

\begin{figure}[pos=!t]
\centering
\includegraphics[width=0.8\textwidth]{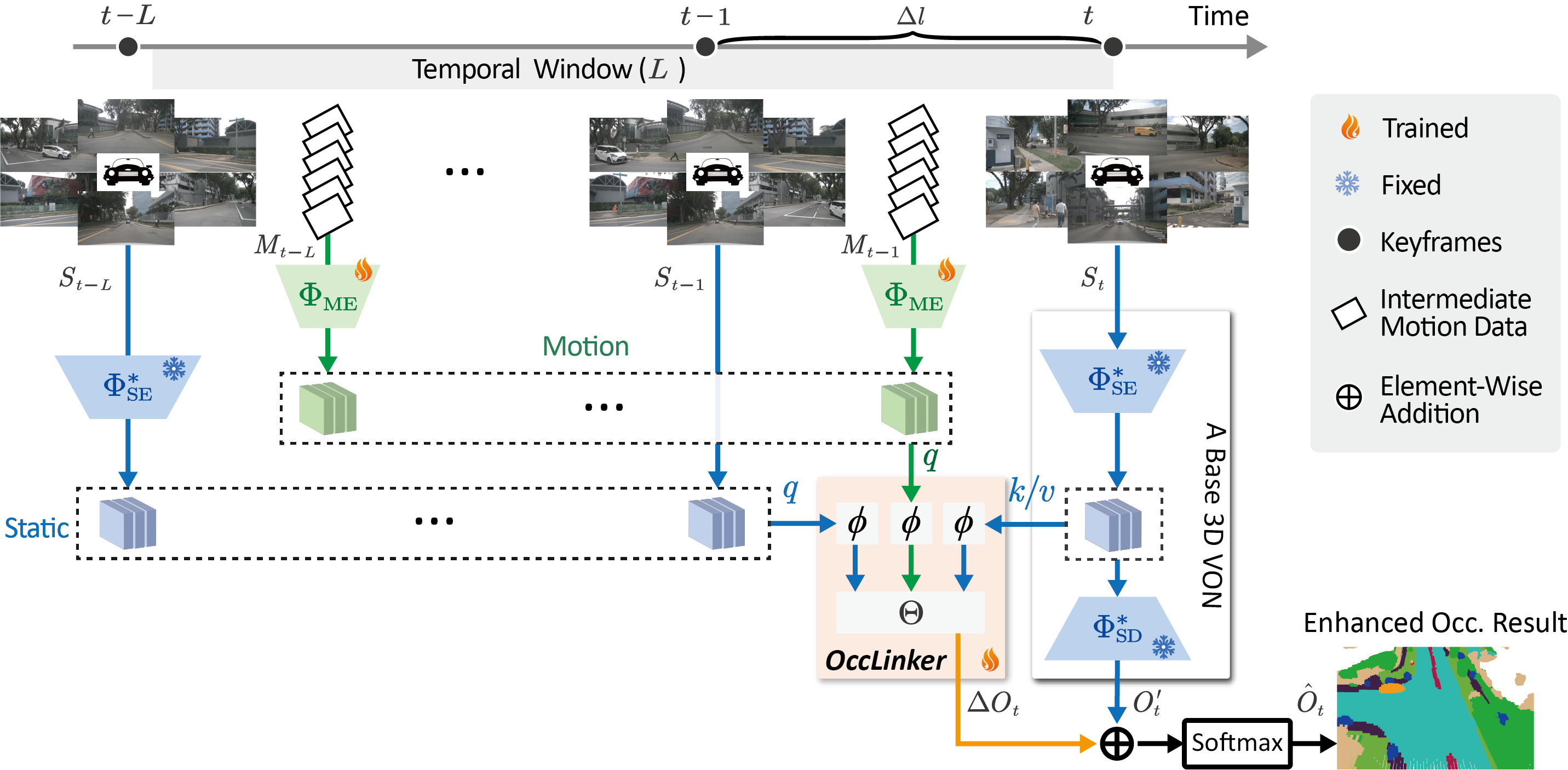}
\caption{
Overview of \ours. A frozen base 3D VON~\cite{surroundOcc,MonoScene,viewformer} runs at a fixed frequency to extract static texture features by $\Phi_{\rm SE}^*$ and generate initial occupancy prediction by $\Phi_{\rm SD}^*$. We employ $\Phi_{\rm ME}$ for extracting motion features of intermediate frame differences, and \ours uses the static and motion features from the recent temporal window as queries, and the current static feature as the key and value. Through lightweight tokenization encoder $\phi$ and dual cross-attention module $\Theta$, it constructs spatio-temporal correlations and outputs a correction term $\Delta O_t$ to refine the initial prediction.}
\label{fig:overview}
\end{figure}
As illustrated in \figurename~\ref{fig:overview}, given a temporal window $\mathcal{W}$ of length $L$, the input includes a sequence of $L$ historical keyframes denoted by $S_{t-L:t-1}$, one current keyframe $S_t$, and a corresponding sequence of $L$ motion cues $M_{t-L:t-1}$, where each motion cue represents frame differences over a short interval $\Delta l$. Let $|V|$ denote the number of cameras. For simplicity, we merge all multi-view inputs into a unified notation and do not distinguish between individual camera views. Both RGB and motion frames are assumed to have the same spatial resolution.

Typically, a base 3D VON $\Phi$ first encodes the current keyframe $S_t$ using a 2D encoder $\Phi_{{\rm SE}}$ to extract static appearance features $s_t$. These features are then lifted into 3D space via a 2D-to-3D decoder $\Phi_{{\rm SD}}$, resulting in an initial occupancy prediction $O_t'$:
\begin{equation}
s_t=\Phi_{{\rm SE}}^*(S_t),\quad O_t' = \Phi_{{\rm SD}}^*(s_t),
\label{eq:base}
\end{equation}
where $\cdot^*$ indicates frozen weights from the pre-trained foundational network. Since the base VON operates frame by frame, historical static features $s_{t-L:t}$ extracted by the frozen static encoder $\Phi_{\rm SE}^*$ are pre-stored and reused for temporal correlation modeling.

Our goal is to design a lightweight plug-in module, parameterized by $\Theta$, that associates the motion features within a temporal window of length $L$ with the static features $s_{t-L:t}$ extracted by the base VON, thereby enhancing the foundational prediction $O_t'$ at each timestep $t$. To achieve this, the intermediate frame differences, denoted by $M_{t-L:t-1}$, are first encoded into a latent representation using a motion encoder $\Phi_{\rm ME}$, yielding temporal motion features:
\begin{equation}
m_{t-L:t-1} = \Phi_{\rm ME}(M_{t-L:t-1}).
\end{equation}

To enable efficient feature linking, we employ a lightweight tokenization encoder $\phi$ that transforms both the static features $s_{t-L:t}$ and the motion features $m_{t-L:t-1}$ into a compact sparse token set. Subsequently, the cross-attention module $\Theta$ establishes spatio-temporal correlations within this latent token space, enhancing prediction smoothness and accuracy while compensating for missing voxels in the current occupancy estimate. The resulting enhancement component $\Delta O_t$ is formulated as:
\begin{equation}
\Delta O_t = \Theta(
    \phi(s_{t-L:t}), \phi(m_{t-L:t-1})
).
\end{equation}

Then, the final enhanced occupancy prediction becomes:
\begin{equation}
\hat{O_t} = {\rm Softmax}(O_t' + \Delta O_t).
\label{eq:final}
\end{equation}

Here, $O_t'$ and $\Delta O_t$ denote the occupancy logits, each shaped $C\times H\times W\times D$, where $C$ is the number of semantic occupancy classes, and $H$, $W$, and $D$ denote the height, width, and depth (length) of the 3D occupancy data, respectively. The ${\rm Softmax}(\cdot)$ function is applied along the class dimension to convert the logits into class probabilities.

Notably, our method focuses on enhancing the current-frame occupancy using historical and motion cues, rather than forecasting future occupancy. The design principle emphasizes lightweight integration without retraining or modifying the base network. \figurename~\ref{fig:overview} presents the overall design of our \ours. In contrast to previous history-aware approaches such as LinkOcc~\cite{ouyang2024linkocc}, ViewFormer~\cite{viewformer}, BEVDet4D-Occ~\cite{bevdet4d} and OPUS~\cite{opus} that depend on redundant historical data, \ours computes the enhancement component $\Delta O_t$ using a dual cross-attention mechanism~\cite{attention_is_all_you_need}, designed to distill essential cues from both static and motion features while capturing compact latent correlations.


\begin{algorithm}[!t]
\caption{\ours inference pipeline}
\label{alg:occlinker}
\KwIn{Keyframes $S_{t-L:t}$ and motion cues $M_{t-L:t}$; Frozen static encoder $\Phi_{\rm SE}^*$ and decoder $\Phi_{\rm SD}^*$ from the base VON; Trainable motion encoder $\Phi_{\rm ME}$ and trainable components in \ours, including tokenization encoder $\phi$ and dual cross-attention module $\Theta$.}
\KwOut{Enhanced occupancy prediction $\hat{O}_t$.}
Initialize temporal context window $\mathcal{W}\gets\emptyset$, $|\mathcal{W}|=L$\;
\While{\KwTrue}{
    Extract static features from the current frame: $s_t \gets \Phi_{\rm SE}^*(S_t)$\;
    Predict initial occupancy logits: $O_t' \gets \Phi_{\rm SD}^*(s_t)$\;
    Extract motion features from the high-frequency motion cues: $m_t \gets \Phi_{\rm ME}(M_t)$\;
    Append $(s_t, m_t)$ to the temporal window $\mathcal{W}$ and discard the oldest entry if $|\mathcal{W}|>L$\;
    Retrieve recent static and motion features $s_{t-L:t-1}$, $m_{t-L:t-1}$, and $s_t$ from $\mathcal{W}$\;
    Convert all retrieved features into sparse tokens via Eq.~\eqref{eq:tokenize}\;
    Establish spatio-temporal correlations within these tokens via Eq.~\eqref{eq:crossattn}\;
    Decode the attention outputs into a correction component via Eq.~\eqref{eq:correction}\;
    Refine the initial occupancy prediction via Eq.~\eqref{eq:final}\;
    Advance to the next timestep: $t\gets t+1$\;
}
\end{algorithm}

\subsection{Sparse tokenization for query construction}
\label{sec:meth:token}

To effectively extract static and motion cues for downstream correlation modeling, we first encode the raw inputs into compact token representations. This process is composed of two parallel modules: a static feature encoder (SE) and a motion feature encoder (ME), followed by a lightweight tokenization encoder (TE).

\noindent\textit{Static feature encoder.}
We extract static spatial features from both the current and previous keyframes by directly reusing the 2D vision backbone modules, denoted as $\Phi_{\rm SE}$, from foundational vision-based occupancy networks (e.g., SurroundOcc~\cite{surroundOcc} or ViewFormer~\cite{viewformer}). Let $S_t$ denote the current surround-view image and $S_{t-L:t-1}$ denote the historical keyframes within a temporal window of size $L$. The corresponding static features are computed as:
\begin{equation}
s_{t-L:t} = \Phi_{\rm SE}^*(S_{t-L:t}),
\end{equation}
where $\Phi_{\rm SE}^*$ shares parameters across all time steps.

\noindent\textit{Motion feature encoder.}
Although occupancy prediction is typically performed using keyframes sampled at a fixed rate, the ego-vehicle continuously captures high-frequency RGB images. These intermediate frames, located between consecutive keyframes, contain rich motion cues that are valuable for spatial-temporal perception. To efficiently extract motion information, we adopt a simple yet effective frame-difference technique~\cite{Frame_difference}. For each pair of adjacent keyframes $(S_i, S_{i+1})$ within the historical window $[t-L, t]$, where $t-L \leq i < i+1 \leq t$, we compute the difference between any two intermediate frames indexed by $(\tau_a, \tau_b)$ satisfying $i \leq \tau_a < \tau_b \leq i+1$:
\begin{equation}
    F_{i:i+1}=\{S^{\tau_b}_{i:i+1}-S^{\tau_a}_{i:i+1}\}.
\end{equation}

All such differences across the $L$ intervals are concatenated as the motion input:
\begin{equation}
M_{t-L:t-1} = \text{Concat}\left( \left\{ F_{i:i+1} \mid i \in [t-L, t-1] \right\} \right),
\end{equation}
where $\text{Concat}(\cdot)$ denotes the concatenation operation performed along the temporal dimension, combining motion cues from all intervals in the historical window.

The resulting difference images $M_{t-L:t}$ are then used as input to the motion encoding module. Since these frame difference images typically share the same resolution as the original RGB images -- often high-resolution~\cite{nuScenes} (e.g., $1600 \times 900$) -- direct processing can be expensive and unsuitable for real-time applications. To address this, we first downsample $M_{t-L:t-1}$ by a factor of five to reduce the spatial dimensions. Moreover, unlike RGB images that contain complex textures and color variations, frame difference images primarily capture motion boundaries. Therefore, we employ a lightweight three-layer convolutional neural network (CNN) followed by batch normalization (BN) to efficiently extract motion cues. This network is denoted by $\Phi_{\rm ME}$. Formally, the resulting motion features are computed as:
\begin{equation}
    m_{t-L:t-1} = \Phi_{\rm ME}(M_{t-L:t-1}).
\end{equation}

\noindent\textit{Sparse tokenization encoder.}
The extracted features $s_t$, $s_{t-L:t-1}$, and $m_{t-L:t-1}$ are then fed into a shared tokenization encoder that transforms them into compact token sets. Specifically, a $1{\times}1$ convolution first reduces channel dimensions, followed by a spatial unfolding operation that divides each image into $p{\times}p$ patches. Within each patch, pixel-wise features are averaged to form a $v$-dimensional vector. This process is applied to each camera view independently and results in the following tokenization operation:
\begin{equation}
\phi(I) = \overline{\mathcal{U}(p, p)(I \circledast\Theta_{1\times1})},
\label{eq:tokenize}
\end{equation}
where $I$ is the input feature, $\circledast$ denotes convolution, $\Theta_{1\times1}$ is a $1{\times}1$ kernel, and $\mathcal{U}(p, p)$ represents patch unfolding. The averaging operator $\overline{\cdot}$ produces a compact representation per spatial patch.

The final output $E$ is a set of sparse tokens with shape $[|V|\cdot\lfloor\frac{h}{p}\rfloor\cdot\lfloor\frac{w}{p}\rfloor, 1, d_{token}]$, where each token retains localized spatial or motion semantics. This tokenized format is especially well-suited for attention-based fusion in our multi-stream integration module, as discussed in Section~\ref{sec:meth:cor}.

\subsection{Spatio-temporal correlation strategy}
\label{sec:meth:cor}

To exploit historical static and motion cues while avoiding redundant computation, we adopt a spatio-temporal correlation strategy based on dual multi-head attention. Unlike previous approaches that directly fuse all temporal inputs indiscriminately~\cite{bevdet4d}, our method disentangles the correlation modeling into two independent streams: one for static context and another for motion cues. These correlations are computed in a shared compact latent space, improving both interpretability and efficiency.

\noindent\textit{Cross-attention-based spatio-temporal reasoning.}
Given the sparse tokens $s_t$, $s_{t-L:t-1}$, and $m_{t-L:t-1}$ extracted in Section~\ref{sec:meth:token}, we use $s_t$ (the current frame token) as the reference context. The goal is to model how past appearance and motion features relate to the current frame through dual multi-head attention mechanisms (MHAM)~\cite{attention_is_all_you_need}.

\begin{figure}[pos=!htbp]
\centering
\includegraphics[width=0.50\textwidth]{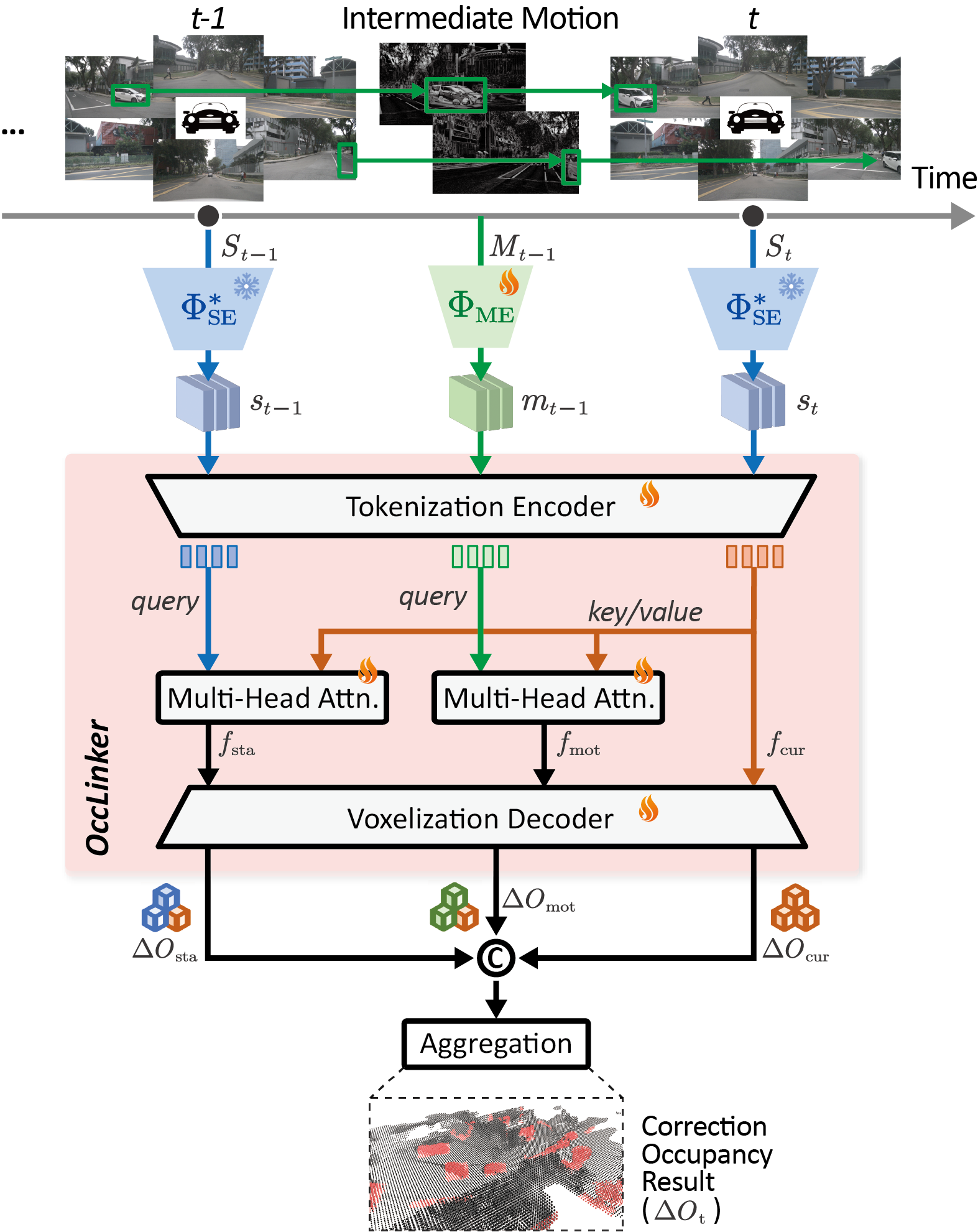}
\caption{Correlation pipeline in \ours. Using $L=1$ as an example, \ours explicitly separates static and dynamic correlations by treating current static features as keys and values, while using historical static features and intermediate motion features as queries.} 
\label{fig:cor}
\end{figure}

\begin{itemize}
\item Historical static attention: 
We use the tokenized historical static features $\phi(s_{t-L:t-1})$ as queries ($Q_{\rm sta}$), while the tokenized current feature $\phi(s_t)$ serves as both keys ($K_{\rm cur}$) and values ($V_{\rm cur}$). Each component is projected into a shared latent space via independent learnable transformations. This MHAM produces the output $f_{\rm sta}$, capturing long-range spatial appearance correlations, as shown in the first column of \figurename~\ref{fig:cor}.

\item Historical motion attention: 
Similarly, the motion tokens $\phi(m_{t-L:t-1})$ are used as queries ($Q_{\rm mot}$), and the same current feature tokens $\phi(s_t)$ are reused as keys ($K_{\rm cur}$) and values ($V_{\rm cur}$). The resulting attention output $f_{\rm mot}$ captures temporally aligned motion dynamics, as shown in the second column of \figurename~\ref{fig:cor}.
\end{itemize}

Each attention stream is formulated as:
\begin{equation}
\begin{aligned}
f_{\rm sta} &= \mathrm{Softmax}
\left(
    \frac{W_{\rm sta}^{Q} Q_{\rm sta} \left( W_{\rm cur}^{K} K_{\rm cur} \right)^{\top}}{\sqrt{d_{\rm grad}}}
\right)
\left( W_{\rm cur}^{V} V_{\rm cur} \right), \\
f_{\rm mot} &= \mathrm{Softmax}
\left(
    \frac{W_{\rm mot}^{Q} Q_{\rm mot} \left( W_{\rm cur}^{K} K_{\rm cur} \right)^{\top}}{\sqrt{d_{\rm grad}}}
\right)
\left( W_{\rm cur}^{V} V_{\rm cur} \right),
\end{aligned}
\label{eq:crossattn}
\end{equation}
where $W^{Q}_*$, $W^{K}_*$, and $W^{V}_*$ are independent learnable projection matrices for each stream. The $\mathrm{Softmax}(\cdot)$ function is used to normalize the attention scores, and the scaling factor $\sqrt{d_{\rm grad}}$ helps stabilize gradients during training~\cite{attention_is_all_you_need}.

\noindent\textit{Feature decoding and correction.}
The attention outputs $f_{\rm sta}$ and $f_{\rm mot}$, together with the current token $f_{\rm cur}=K_{\rm cur}$, are passed through separate lightweight decoding heads to reconstruct the spatio-temporal correction signal $\Delta O_t$. Each stream is first reshaped using 3D deconvolution layers followed by ReLU activations to recover the spatial dimensions. The decoding network is parameterized by $\Theta_{\rm D}$. This produces three correction results: $\Delta O_{\rm sta}$ from static attention, $\Delta O_{\rm mot}$ from motion attention, and $\Delta O_{\rm cur}$ from the current frame as the baseline. These results are then concatenated along the feature channel and aggregated using an additional 3D convolutional layer:
\begin{equation}
\Delta O_t = {\rm Concat}(
    \{\Delta O_{\rm sta}, \Delta O_{\rm mot}, \Delta O_{\rm cur}\}
) \circledast \Theta_{3\times3\times3},
\label{eq:correction}
\end{equation}
where $\Theta_{3\times3\times3}$ is a learnable 3D convolution kernel.

The fused correction $\Delta O_t$ is then added to the initial occupancy logits $O_t'$ from the base model, as defined in Eq.~\eqref{eq:final}, yielding the final enhanced prediction $\hat{O}_t$. This integration augments the base prediction with temporally aligned static and motion-aware cues, while maintaining full compatibility with any existing foundational VONs.

The proposed correlation design improves both interpretability and efficiency by explicitly separating static and dynamic contexts, enabling robust enhancement without modifying the backbone network.

\rev{Unlike methods that explicitly warp historical features using ego-motion information~\cite{xu2025exploiting,viewformer}, \ours uses dual cross-attention to learn associations between historical static and motion cues and current features in a sparse latent space. This implicit alignment requires no pose input, motion supervision, or coordinate transformation, keeping \ours lightweight, sensor-agnostic, and compatible with both single-frame and history-aware VONs.}


\subsection{Optimization}
\label{sec:meth:opt}

To maintain stability and efficiency during training, we freeze the static encoder $\Phi_{\rm SE}$ and only optimize the proposed enhancement module \ours along with the motion encoder $\Phi_{\rm ME}$. This design yields two key advantages:
(1) it avoids potential gradient interference in $\Phi_{\rm SE}$, ensuring more stable optimization; and
(2) it accelerates convergence of our plug-in module by leveraging reliable static features extracted by the pre-trained backbone.

To ensure a fair comparison and retain consistency with prior work, we adopt the original loss functions used by each benchmark during training.
For the SurroundOcc~\cite{surroundOcc} benchmark, we employ the following composite loss function: 
\begin{equation}
\mathcal{L} = \mathcal{L}_{\rm ce} + \mathcal{L}_{\rm sem} + \mathcal{L}_{\rm geo},
\end{equation}
where $\mathcal{L}_{\rm ce}$ is a voxel-wise cross-entropy loss for occupancy classification, while $\mathcal{L}_{\rm sem}$ and $\mathcal{L}_{\rm geo}$ are the semantic and geometric variants of the scene-class affinity loss introduced in MonoScene~\cite{MonoScene}.

For the Occ3D~\cite{Occ3D} benchmark, following the convention in ViewFormer~\cite{viewformer}, we apply the following loss:
\begin{equation}
\mathcal{L} = \mathcal{L}_{\rm focal} + \mathcal{L}_{\rm ce} + \mathcal{L}_{\rm ls} + \lambda\mathcal{L}_{\rm l1},
\end{equation}
where $\mathcal{L}_{\rm focal}$ is the focal loss~\cite{focal_loss} for handling class imbalance, $\mathcal{L}_{\rm ls}$ is the lovasz softmax loss~\cite{lovasz_loss} for improving boundary-level segmentation, $\mathcal{L}_{\rm l1}$ is an L1 regression loss (scaled by $\lambda$) to supervise motion-induced voxel correction.

\subsection{\rev{Theoretical Analysis}}
\label{sec:meth:theory}

\noindent\rev{\textit{Information-theoretic analysis.}}
\rev{%
Let $Y_t$ denote the one-hot ground-truth occupancy, and define the residual of the frozen base prediction as
\begin{equation}
R_t=Y_t-{\rm Softmax}(O'_t).
\label{eq:residual}
\end{equation}
We use mutual information theory~\cite{poole2019variational} to analyze the properties of our \ours.

\begin{lemma}
\label{lem:mi_equivalence}
Given the frozen base logits $O'_t$, the residual $R_t$ and the ground truth $Y_t$ contain the same information about the learned correction $\Delta O_t$:
\begin{equation}
I(R_t;\Delta O_t\mid O'_t)
=
I(Y_t;\Delta O_t\mid O'_t).
\label{eq:mi_equivalence}
\end{equation}
\end{lemma}

\begin{proof}
For a fixed $O'_t$, the mapping from $Y_t$ to $R_t$ is one-to-one because
\begin{equation}
Y_t=R_t+{\rm Softmax}(O'_t).
\end{equation}
Therefore, conditioned on $O'_t$, $Y_t$ and $R_t$ can be uniquely recovered from each other. Conditional mutual information is invariant under such a bijective transformation, which proves Eq.~\eqref{eq:mi_equivalence}.
\end{proof}

\begin{theorem}
\label{thm:mi_lower_bound}
For the enhanced prediction
$\hat O_t={\rm Softmax}(O'_t+\Delta O_t)$, the conditional mutual information between the base residual and the learned correction satisfies
\begin{equation}
\begin{aligned}
I(R_t;\Delta O_t\mid O'_t)
&\geq H(Y_t\mid O'_t)-\mathcal L_{\rm ce} \\
&\geq H(Y_t\mid O'_t)-\mathcal L,
\end{aligned}
\label{eq:mi_loss_bound}
\end{equation}
where $\mathcal L$ is the total training loss defined above.
\end{theorem}

\begin{proof}
According to Lemma~\ref{lem:mi_equivalence},
\begin{equation}
\begin{aligned}
I(R_t;\Delta O_t\mid O'_t)
&=I(Y_t;\Delta O_t\mid O'_t)\\
&=H(Y_t\mid O'_t)
-H(Y_t\mid O'_t,\Delta O_t).
\end{aligned}
\label{eq:mi_decomposition}
\end{equation}
The voxel-wise cross-entropy loss can be decomposed as
\begin{equation}
\begin{aligned}
\mathcal L_{\rm ce}
={}&H(Y_t\mid O'_t,\Delta O_t)\\
&+\mathbb E\!\left[
D_{\rm KL}\!\left(
p(\cdot\mid O'_t,\Delta O_t)
\,\Vert\,\hat O_t
\right)
\right].
\end{aligned}
\label{eq:ce_decomposition}
\end{equation}
Since the KL divergence is non-negative,
\begin{equation}
H(Y_t\mid O'_t,\Delta O_t)\leq\mathcal L_{\rm ce}.
\end{equation}
Substituting this inequality into Eq.~\eqref{eq:mi_decomposition} gives the first inequality in Eq.~\eqref{eq:mi_loss_bound}. Moreover, all auxiliary loss terms are non-negative, and therefore
$\mathcal L_{\rm ce}\leq\mathcal L$.

Because $\Phi_{\rm SE}^{*}$ and $\Phi_{\rm SD}^{*}$ are frozen, $O'_t$ remains fixed during plug-in training. Therefore, $H(Y_t\mid O'_t)$ is a constant. Let $k$ denote the optimization step, if
$\mathcal L^{(k+1)}\leq\mathcal L^{(k)}$, then
\begin{equation}
\begin{aligned}
&\left[H(Y_t\mid O'_t)-\mathcal L^{(k+1)}\right]
-\left[H(Y_t\mid O'_t)-\mathcal L^{(k)}\right]\\
&\qquad=
\mathcal L^{(k)}-\mathcal L^{(k+1)}
\geq0.
\end{aligned}
\label{eq:lower_bound_improvement}
\end{equation}
Thus, decreasing the training loss strengthens the lower bound on
$I(R_t;\Delta O_t\mid O'_t)$.
\end{proof}}

\section{Experiments}
\label{sec:exp}
\subsection{Experimental settings}

\noindent\textit{Benchmark.}
We evaluate on two established occupancy benchmarks: (i) the SurroundOcc benchmark~\cite{surroundOcc,nuScenes}, which generates occupancy labels by voxelizing aggregated multi-frame LiDAR in the ego frame, using manually annotated 3D bounding boxes to assign semantic categories; and (ii) the Occ3D benchmark~\cite{Occ3D}, which provides 0.4m-resolution voxel-level labels with occlusion states through automated LiDAR aggregation and mesh reconstruction. Both benchmarks cover 1,050 driving scenes, each with up to 40 timestamped frames and six synchronized cameras (front, front-left/right, back, back-left/right) at 1600$\times$900 resolution. In our experiments, we enhance single-frame baselines~\cite{MonoScene,surroundOcc,viewformer} by aggregating features from $L$ historical keyframes. Additionally, unlabeled intermediate frames from the ``sweeps'' folder~\cite{nuScenes} are utilized as motion information for \ours.

\noindent\textit{Implementation details.}
Both the SurroundOcc and Occ3D benchmarks adopt the same 3D occupancy resolution of $(H,W,D)=(200,200,16)$ and utilize the same number of surround-view cameras, with $|V|=6$. However, they differ in semantic granularity: SurroundOcc defines $C=17$ semantic classes, while Occ3D provides $C=18$ for finer-grained occupancy labeling. Following the respective conventions~\cite{surroundOcc,viewformer} for static feature dimensions, we adopt a feature map resolution of $(h,w)=(116,200)$ for SurroundOcc and $(h,w)=(32,88)$ for Occ3D. Notably, both $s_{t-L:t}$ and $m_{t-L:t}$ share the same dimensionality of $|V|\times c\times w\times h$.

We employ the $\Phi_{\rm ME}$ module with a temporal sampling interval of $\Delta l\approx 0.1$ seconds to extract motion features from raw nuScenes images at a resolution of $900 \times 1600$. Within the $\Phi_{\rm ME}$ module, input images are first downsampled by a factor of 5, resulting in $180 \times 320$ feature maps to reduce GPU memory consumption. Motion inputs are computed using either frame difference operations~\cite{Frame_difference} or optical flow analysis~\cite{OpticalFlow}. The token dimension and the gradient scaling factor are both set to $d_{token}=d_{grad}=32$, with the number of temporal patches set as $p=6$. For the convolutional layer $\Theta_{1\times1}$, the kernel size, stride, padding, input channels, and output channels are set to $(1,1)$, $1$, $0$, 512, and $d_{token}$, respectively. In contrast, for $\Theta_{3\times3\times3}$, these values are set to $(3, 3, 3)$, $1$, $1$, $3C$, and $C$, respectively. OccLinker is integrated on top of this baseline to complement the missing inter-frame motion cues, rather than replacing any component of the original design.

All experiments were conducted using four NVIDIA L20 GPUs with a batch size of 1. We employed the AdamW optimizer on both the SurroundOcc~\cite{surroundOcc} and Occ3D~\cite{Occ3D} benchmarks, setting the learning rate to $2\times10^{-4}$, weight decay to $1\times10^{-2}$, and betas to $(0.9, 0.999)$. The overall training took approximately 24 epochs, corresponding to about 1.5 GPU-days in total, including around 1.4 GPU-days for pretraining the base VON and 0.1 GPU-day for fine-tuning \ours.


\subsection{Evaluation Metrics}

\noindent\textit{Occupancy accuracy metric.}
For a rigorous and comprehensive evaluation across benchmarks, we employ Intersection over Union (IoU) and Mean Intersection over Union (mIoU) metrics, which are standard in 3D semantic occupancy prediction tasks~\cite{PASCAL,Microsoft_COCO}. IoU measures the global overlap between all predicted occupied voxels and their ground-truth counterparts, regardless of semantic categories. It reflects the overall spatial accuracy of the 3D occupancy prediction. In contrast, mIoU is the mean IoU computed by averaging IoUs over all semantic classes, which reflects class-balanced semantic accuracy.

To further evaluate the temporal consistency and occupancy accuracy for both moving and static objects, we adopt the object categorization scheme from Cam4DOcc~\cite{Cam4docc}, dividing objects into two broad classes: General Moving Objects (GMO) and General Static Objects (GSO). Objects that remain stationary relative to the ground are classified as GSO, whereas those that move relative to the ground belong to the GMO category. This classification scheme enables a fine-grained and quantitative analysis of moving and static objects within the scene. The mIoU scores are reported separately for three groups: all classes, GMO classes, and GSO classes.



\noindent\textit{Temporal consistency metric.}
To quantify the temporal stability gained by integrating \ours with baseline models, we introduce a temporal consistency metric designed to detect and measure changes between consecutive frames. This metric reflects the smoothness and stability of predictions, which is crucial for enhancing the user's visual experience. Let $\sigma_{i,n}^{(x,y,z)}$ denote the semantic label of the $n$-th voxel at spatial coordinates $(x,y,z)$ in frame $i$. Define the indicator function $\delta(e_1,e_2)=\mathbb{I}(e_1\neq e_2)$, which equals 1 if the two inputs differ and 0 otherwise. Voxel changes between frames $i$ and $j$ are divided into two categories. A change is classified as Moving Object Change'' (MOC) if $\sigma_{i,n}^{(x,y,z)}\in\text{GMO}$ or $\sigma_{j,n}^{(x,y,z)}\in\text{GMO}$, and as Static Object Change'' (SOC) if $\sigma_{i,n}^{(x,y,z)}\in\text{GSO}$ and $\sigma_{j,n}^{(x,y,z)}\in\text{GSO}$.



Using these definitions, we define disparity metrics $\Delta_{m}$ and $\Delta_{s}$ to quantify temporal inconsistencies between frames $i$ and $j$, as follows:
\begin{equation}
\Delta_{m}(i,j) = \dfrac{1}{N_{\rm mc}} \sum\limits_{n=1}^{N_{\rm mc}} \delta\left(\sigma_{i,n}^{(x,y,z)}, \sigma_{j,n}^{(x,y,z)}\right), 
\quad \
\Delta_{s}(i,j) = \dfrac{1}{N_{\rm sc}} \sum\limits_{n=1}^{N_{\rm sc}} \delta\left(\sigma_{i,n}^{(x,y,z)}, \sigma_{j,n}^{(x,y,z)}\right).
\label{eq:disparity}
\end{equation}
where $N_{\rm mc}$ and $N_{\rm sc}$ represent the number of corresponding voxels in the moving and static regions, respectively.

The overall temporal consistency scores for moving and static objects, $S_{\rm m}$ and $S_{\rm s}$, are then computed by aggregating disparities over sequential frames:
\begin{equation}
S_{\rm m/s} = 1 - \dfrac{1}{M-1} \sum\limits_{k=1}^{M-1} \Delta_{\rm m/s}(k,k+1),
\label{eq:consistency_scores}
\end{equation}
where $M$ denotes the total number of frames in the scene. Final consistency scores $\overline{S_{\rm m}}$ and $\overline{S_{\rm s}}$ are averaged across all scenes. Higher values indicate smoother and more temporally consistent occupancy predictions. \rev{Both scores reward similarity between consecutive volumes, so they can be raised by any procedure that makes the output change less, including ones that damage accuracy. We therefore read them together with IoU and mIoU, and we give the evidence for this caution in Section~\ref{sec:exp}.}

\subsection{Comparison Results}
\noindent\textit{Occupancy accuracy on SurroundOcc benchmark.}
We compare our method with several state-of-the-art models, including single-frame VON methods such as Atlas~\cite{Atlas}, TPVFormer~\cite{TPVFormer}, MonoScene~\cite{MonoScene}, and SurroundOcc~\cite{surroundOcc}, as well as history-aware VON methods such as BEVFormer~\cite{BEVFormer} and BEVDet4D-Occ~\cite{bevdet4d}. To ensure a fair comparison, all baselines and \ours are trained on the same ground truth data using identical training protocols. \rev{The column $L$ lists the number of historical keyframes each method consumes, so a single-frame method has $L=0$. The plug-in rows inherit $L$ from their base network.} As shown in \tablename~\ref{tab:main-res-a}, integrating \ours into existing baselines like MonoScene~\cite{MonoScene} and SurroundOcc~\cite{surroundOcc} leads to notable performance gains. In particular, the combination of \ours with SurroundOcc achieves the best results among all models, improving IoU and mIoU (All) by 1.63\% and 0.37\%, respectively. Moreover, compared with BEVFormer, which incorporates historical BEV features, our method further boosts performance by 0.41\% in IoU and 0.34\% in mIoU.

\begin{table}[pos=!ht]
\centering
\setlength{\tabcolsep}{3pt}
\caption{Occupancy prediction performance on the SurroundOcc benchmark~\cite{surroundOcc}. Performance gains ($\Delta$) are highlighted in \textbf{bold}, and the best result in each column is marked by an asterisk. \rev{Frame counts follow the convention defined in the text.}
}
\label{tab:main-res-a}
\begin{tabular}{ RCCCCCCCC }
\toprule
    \multicolumn{1}{r}{\multirow{2}{*}[-0.4em]{Method}} & \multicolumn{1}{c}{\multirow{2}{*}[-0.4em]{\rev{$L$}}} & \multicolumn{1}{c}{\multirow{2}{*}[-0.4em]{\rev{Backbone}}} & \multicolumn{1}{c}{\multirow{2}{*}[-0.4em]{IoU~$\uparrow$}} & \multicolumn{3}{c}{mIoU~$\uparrow$} & \multicolumn{1}{c}{\multirow{2}{*}[-0.4em]{$\overline{S_{\rm m}}\uparrow$}} & \multicolumn{1}{c}{\multirow{2}{*}[-0.4em]{$\overline{S_{\rm s}}\uparrow$}} \\ 
\cmidrule(lr){5-7}
    \multicolumn{1}{c}{} & \multicolumn{1}{c}{} & \multicolumn{1}{c}{} & \multicolumn{1}{c}{} & All & GMO & GSO & & \\ 
\midrule
    \multicolumn{9}{c}{Single-frame VON methods} \\
\midrule        
    Atlas~\cite{Atlas} & \rev{0} & \rev{R101-DCN} & 28.66 & 15.00 & 12.64 & 17.35 & -- & --  \\
    TPVFormer~\cite{TPVFormer} & \rev{0} & \rev{R101-DCN} & 30.86 & 17.10 & 14.04 & 20.15 & -- & -- \\
    MonoScene~\cite{MonoScene} & \rev{0} & \rev{R101-DCN} & 10.04 & 1.15 & 0.24 & 2.07 & 46.53 & 81.77 \\
    \rev{\quad + CVT-Occ~\cite{ye2024cvt}} & --- & --- & \rev{11.05} & \rev{1.90} & \rev{0.13} & \rev{3.67} & \rev{48.14} & \rev{82.93} \\
    \makecell[r]{+\ours\\$\Delta$} & --- & --- & \makecell[c]{13.10\\\textbf{+3.06}} & \makecell{1.69\\\textbf{+0.54}}  & \makecell[c]{0.34\\\textbf{+0.10}} & \makecell{3.04\\\textbf{+0.97}} & \makecell[c]{54.21\\\textbf{+7.68}} & \makecell{83.84\\\textbf{+2.07}} \\
    SurroundOcc~\cite{surroundOcc} & \rev{0} & \rev{R101-DCN} & 31.49 & 20.30  & 18.39 & 22.20 & 58.33 & 91.71 \\
    \rev{\quad + CVT-Occ} & --- & --- & \rev{32.18} & \rev{20.39} & \rev{18.70$^{*}$} & \rev{22.08} & \rev{60.41} & \rev{93.06$^{*}$} \\
    \makecell[r]{+\ours\\$\Delta$} & --- & --- & \makecell[c]{33.12$^{*}$\\\textbf{+1.63}} & \makecell[c]{20.67$^{*}$\\\textbf{+0.37}} & \makecell[c]{18.26\\-0.13} & \makecell[c]{23.08$^{*}$\\\textbf{+0.88}} & \makecell[c]{60.64$^{*}$\\\textbf{+2.31}} & \makecell[c]{92.54\\\textbf{+0.83}} \\
\midrule
    \multicolumn{9}{c}{History-aware VON methods} \\
\midrule
    BEVDet4D-Occ~\cite{bevdet4d} & \rev{1} & \rev{R101-DCN} & 24.26 & 14.22 & 11.10 & 17.34 & -- & -- \\
    LinkOcc~\cite{ouyang2024linkocc} & \rev{1} & \rev{R101-DCN} & 30.01 & 16.78 & 12.03 & 21.53 & 60.26 & 83.23 \\
    BEVFormer~\cite{BEVFormer} & \rev{3} & \rev{R101-DCN} & 30.50 & 16.75 & 14.17 & 19.33 & 52.84 & 84.21  \\
    \rev{\quad + CVT-Occ} & --- & --- & \rev{30.12} & \rev{16.99} & \rev{14.24} & \rev{19.74} & \rev{55.51} & \rev{85.01} \\
    \makecell[r]{+\ours\\$\Delta$} & --- & --- & \makecell{30.91\\\textbf{+0.41}} & \makecell{17.09\\\textbf{+0.34}} & \makecell{14.16\\-0.01} & \makecell{20.01\\\textbf{+0.68}} & \makecell{55.21\\\textbf{+2.37}} & \makecell{85.99\\\textbf{+1.78}} \\
\bottomrule
\end{tabular}
\end{table}

\noindent\textit{Occupancy accuracy on Occ3D benchmark.}
We also evaluate our method on the Occ3D benchmark~\cite{Occ3D}, as shown in \tablename~\ref{tab:main-res-b}. The baselines include single-frame VON methods such as MonoScene~\cite{MonoScene}, OccFormer~\cite{OccFormer}, SurroundOcc~\cite{surroundOcc}, and ViewFormer~\cite{viewformer}, as well as history-aware VON methods including SparseOcc~\cite{SparseOcc_Liu}, BEVDet4D-Occ~\cite{bevdet4d}, and OPUS-L~\cite{opus}. When integrated into single-frame VONs, \ours consistently improves the base models. Furthermore, in ViewFormer+\ours, ViewFormer maintains a memory queue in which each frame's intermediate BEV feature interacts with historical multi-frame BEV representations, while our method further supplements the missing sparse motion features between frames. As a result, ViewFormer+\ours surpasses ViewFormer by 0.24\% in IoU and 0.84\% in mIoU (All).

We note that in \tablename~\ref{tab:main-res-b}, ViewFormer achieves higher IoU than SurroundOcc, while this advantage is not reflected in $\overline{S_s}$. This is mainly because ViewFormer attains higher IoU by aggregating multi-frame BEV features, which captures richer geometric context for individual frames. However, its complex temporal fusion may occasionally propagate redundant or noisy historical information, resulting in less stable transitions between consecutive frames. In contrast, SurroundOcc performs frame-by-frame prediction with limited reliance on long-term history, which naturally leads to smoother temporal behavior and thus higher temporal consistency scores, albeit at the cost of lower per-frame spatial accuracy. Notably, our method improves the temporal consistency of both categories of models by explicitly enhancing sparse inter-frame correlations without introducing additional temporal noise.

\begin{table}[pos=!ht]
\setlength{\tabcolsep}{3pt}
\centering
\caption{Occupancy prediction performance on the Occ3D benchmark~\cite{Occ3D}. Performance gains ($\Delta$) are highlighted in \textbf{bold}, and the best result in each column is marked by an asterisk. \rev{Frame counts follow the convention defined in the text. $^{\S}$ LMPOcc-S draws no historical keyframe from the current sequence; its temporal information comes from a global occupancy prior aggregated over earlier traversals of the same location, which the other entries do not use.}}
\label{tab:main-res-b}
\begin{tabular}{ RCCCCCCCC }
\toprule
    \multicolumn{1}{r}{\multirow{2}{*}[-0.4em]{Method}} & \multicolumn{1}{c}{\multirow{2}{*}[-0.4em]{\rev{$L$}}} & \multicolumn{1}{c}{\multirow{2}{*}[-0.4em]{\rev{Backbone}}} & \multicolumn{1}{c}{\multirow{2}{*}[-0.4em]{IoU~$\uparrow$}} & \multicolumn{3}{c}{mIoU~$\uparrow$} & \multicolumn{1}{c}{\multirow{2}{*}[-0.4em]{$\overline{S_{\rm m}}\uparrow$}} & \multicolumn{1}{c}{\multirow{2}{*}[-0.4em]{$\overline{S_{\rm s}}\uparrow$}} \\ 
\cmidrule(lr){5-7}
    \multicolumn{1}{c}{} & \multicolumn{1}{c}{} & \multicolumn{1}{c}{} & \multicolumn{1}{c}{} & All & GMO & GSO & & \\ 
\midrule
    \multicolumn{9}{c}{Single-frame VON methods} \\
\midrule
    MonoScene~\cite{MonoScene} & \rev{0} & \rev{R101} & -- & 6.06 & 5.36 & 6.68 & -- & -- \\
    OccFormer~\cite{OccFormer} & \rev{0} & \rev{R101} & -- & 21.93 & 21.78 & 22.06 & -- & -- \\
    SurroundOcc~\cite{surroundOcc} & \rev{0} & \rev{R101-DCN} & 51.89 & 7.24  & 0.36 & 13.35 & 65.35 & 89.54 \\
    \rev{\quad + CVT-Occ~\cite{ye2024cvt}} & --- & --- & \rev{51.24} & \rev{7.23} & \rev{0.33} & \rev{13.36} & \rev{69.91} & \rev{90.42} \\
    \makecell[r]{+\ours \\$\Delta$} & --- & \rev{R101-DCN} & \makecell{52.13\\\textbf{+0.24}} & \makecell{10.33\\\textbf{+3.09}} & \makecell{1.98\\\textbf{+1.62}} & \makecell{17.76\\\textbf{+4.41}} & \makecell{69.60\\\textbf{+4.25}} & \makecell{90.91$^{*}$\\\textbf{+1.37}} \\
\midrule
    \multicolumn{9}{c}{History-aware VON methods} \\
\midrule
    SparseOcc~\cite{SparseOcc_Liu} & \rev{7} & \rev{R50} & -- & 30.10  & -- & -- & -- & -- \\
    BEVDet4D-Occ~\cite{bevdet4d} & \rev{7} & \rev{R50} & -- & 39.30  & 29.09 & \rev{48.38$^{*}$} & -- & -- \\ 
    OPUS-L~\cite{opus} & \rev{7} & \rev{R50} & -- & 36.20  & 31.25 & \rev{40.60} & -- & -- \\     
    \rev{LMPOcc-S~\cite{yuan2026collaborative}$^{\S}$} & \rev{--} & \rev{R50} & \rev{--} & \rev{40.38} & \rev{33.42} & \rev{46.56} & \rev{--} & \rev{--} \\
    ViewFormer~\cite{viewformer} & \rev{3} & \rev{R50} & 70.39 & 40.46  & 33.73 & 46.45 & 67.26 & 86.06 \\
    \rev{\quad + CVT-Occ} & --- & --- & \rev{70.57} & \rev{40.35} & \rev{34.18} & \rev{45.83} & \rev{69.99} & \rev{86.35} \\
    \makecell[r]{+\ours\\$\Delta$} & --- & \rev{R50} & \makecell{70.63$^{*}$\\\textbf{+0.24}} & \makecell{41.30$^{*}$\\\textbf{+0.84}} & \makecell{34.33$^{*}$\\\textbf{+0.60}} & \makecell{47.50\\\textbf{+1.05}} & \makecell{70.13$^{*}$\\\textbf{+2.87}} & \makecell{87.10\\\textbf{+1.04}} \\
\bottomrule
\end{tabular}
\end{table}

\noindent\rev{\textit{Comparison with other plug-in modules.}}
\rev{Tables~\ref{tab:main-res-a} and \ref{tab:main-res-b} also report CVT-Occ~\cite{ye2024cvt} integrated with the same base networks under the same protocol. Its gains are smaller than those of \ours on most entries, and on ViewFormer, whose streaming memory already carries temporal history, it decreases mIoU by 0.11 while \ours adds 0.84. CVT-Occ fuses historical information on the dense $xy$-plane, whereas our method extracts sparse yet critical information at the token level. Therefore, our method achieves better performance.}

\noindent\textit{Temporal consistency.}
The $\overline{S_{\rm m}}$ and $\overline{S_{\rm s}}$ results presented in Tables~\ref{tab:main-res-a} and \ref{tab:main-res-b} show that integrating \ours consistently improves the temporal consistency of occupancy predictions across all frames and scenes for all models, highlighting the effectiveness of our approach. This improvement can be attributed to the use of previous static features and the incorporation of motion information, which together provide valuable correlations between static/dynamic cues and occupancy refinement.

\subsection{Ablation study}

\noindent\textit{Different combinations of \ours.}
\tablename~\ref{tab:ablation-modules} presents the ablation results of different decomposed correlation designs in \ours, as illustrated in \figurename~\ref{fig:cor}, when integrated with SurroundOcc~\cite{surroundOcc} under $L=1$. \textbf{M0} denotes the baseline model without \ours. \textbf{M1} removes the correlation between historical and current static features. \textbf{M2} omits the current static features entirely. \textbf{M3} excludes the correlation between motion features and current static features. \textbf{M4$^\dag$} reverses the roles of queries and key/value pairs in the correlation design of \figurename~\ref{fig:cor}. \textbf{M4} represents the complete version of \ours with all components included.

The results in \tablename~\ref{tab:ablation-modules} clearly demonstrate that removing any individual stream from the \ours module leads to performance degradation in both prediction accuracy and temporal consistency, confirming the necessity of the decomposed correlation modeling. Moreover, the comparison between \textbf{M4$^\dag$} and \textbf{M4} highlights that treating previous keyframes and motion features as queries -- rather than as keys/values -- effectively captures temporally aligned cues essential for robust occupancy refinement.

\begin{table}[pos=!ht]
\centering
\caption{Ablation study of decomposed latent correlation modeling in \ours when integrated with SurroundOcc. Compared to \textbf{M4}, \textbf{M4$^\dag$} reverses the roles of queries and key/value pairs. Performance gains over \textbf{M0} are \textbf{bolded}, and the best result in each column is marked by an asterisk.}
\label{tab:ablation-modules}
\begin{tabular}{ CCCCCCCC }
\toprule
    Idx. & $\Delta O_{\rm sta}$ & $\Delta O_{\rm cur}$ & $\Delta O_{\rm mot}$ & IoU$\uparrow$ & mIoU$\uparrow$ & $\overline{S_{\rm m}}\uparrow$ & $\overline{S_{\rm s}}\uparrow$ \\
\midrule
    \textbf{M0} & \ding{55} & \ding{55} & \ding{55} & 31.49 & 20.30 & 58.33 & 91.71 \\
    \textbf{M1} & \ding{55} & \ding{51} & \ding{51} & 33.04 & 20.04 & 60.59 & 92.25 \\
    \textbf{M2} & \ding{51} & \ding{55} & \ding{51} & 33.05 & 19.98 & 60.09 & 92.44 \\
    \textbf{M3} & \ding{51} & \ding{51} & \ding{55} & 32.88 & 20.10 & 60.24 & 92.24 \\
    \textbf{M4$^\dag$} & \ding{51} & \ding{51} & \ding{51} & 31.97 & 20.11 & 60.19 & 92.01 \\
\midrule
    \textbf{M4} & \ding{51} & \ding{51} & \ding{51} & 
    \makecell{33.12$^{*}$\\\textbf{+1.63}} & 
    \makecell{20.67$^{*}$\\\textbf{+0.37}} & 
    \makecell{60.64$^{*}$\\\textbf{+2.31}} & 
    \makecell{92.54$^{*}$\\\textbf{+0.83}} \\
\bottomrule
\end{tabular}
\end{table}

\noindent\textit{Impact of different types of motion information.}
\tablename~\ref{tab:ablation-motion} investigates the impact of different types of motion information on SurroundOcc+\ours when $L=1$. Specifically, \textbf{I0} denotes the base model~\cite{surroundOcc} without any motion input. \textbf{I1} uses raw high-frequency intermediate frames as motion input for \ours. \textbf{I2} incorporates optical flow~\cite{OpticalFlow}, while \textbf{I3} employs frame difference~\cite{Frame_difference} to extract motion cues.

\begin{table}[pos=!ht]
\centering
\caption{Ablation study of different motion information types used in \ours. The best result in each column is marked by an asterisk.}
\label{tab:ablation-motion}
\begin{tabular}{ CCCCCCC }
\toprule 
    Idx. & \makecell{Motion type} & IoU~$\uparrow$ & mIoU~$\uparrow$ & $\overline{S_{\rm m}}\uparrow$ & $\overline{S_{\rm s}}\uparrow$ & Time $\downarrow$ \\
\midrule
    \textbf{I0} & - & 31.49 & 20.30 & 58.33 & 91.71 & N/A \\
    \textbf{I1} & Raw Image & 32.39 & 19.45 & 59.01 & 91.15 & N/A \\
    \textbf{I2} & Optical Flow  & 32.80 & 20.27 & 60.53 & 92.13 & 9.221 ms \\
    \textbf{I3} & Frame Diff. & 33.12$^{*}$ & 20.67$^{*}$ & 60.64$^{*}$ & 92.54$^{*}$ & 6.181 ms$^{*}$\\
\bottomrule
\end{tabular}
\end{table}

As shown in \tablename~\ref{tab:ablation-motion}, \textbf{I1} improves IoU over \textbf{I0}, but performs worst in mIoU, $\overline{S_{\rm m}}$, and $\overline{S_{\rm s}}$ among the motion-aware variants. This performance drop is largely due to the presence of redundant and irrelevant information in raw intermediate frames, which complicates the extraction of motion features by \ours. Moreover, \textbf{I3} significantly outperforms \textbf{I2} in both IoU and mIoU, and achieves slightly better results in $\overline{S_{\rm m}}$ and $\overline{S_{\rm s}}$. This suggests that frame difference is more effective at capturing abrupt scene changes, such as pedestrians suddenly appearing or vehicles exiting intersections, whereas optical flow may introduce delays in responding to such dynamics. In addition, thanks to the lightweight design of \ours, using frame difference further reduces computational overhead by operating on simple differential signals, thereby enhancing inference efficiency.

\noindent\textit{Impact of different numbers of previous keyframes.}
We conduct ablation experiments on the temporal context window length $L \in \{0,1,2,3,4\}$ in $\mathcal{W}$. The case of $L=0$ corresponds to SurroundOcc~\cite{surroundOcc}, which does not utilize any historical keyframes. As shown in \tablename~\ref{tab:ntrack}, the best IoU and mIoU are achieved when $L=1$, while the highest $\overline{S_{\rm m}}$ and $\overline{S_{\rm s}}$ scores occur at $L=2$.

\begin{table}[pos=!ht]
\setlength{\tabcolsep}{4pt}
\centering
\caption{Effect of different number of previous keyframes. \rev{The memory was measured in MB on NVIDIA L20 GPUs.} The best result in each column is marked by an asterisk.}
\label{tab:ntrack}
\begin{tabular}{ CCCCCCCC }
\toprule
    Idx. & $L$ & IoU~$\uparrow$ & mIoU~$\uparrow$ & $\overline{S_{\rm m}}\uparrow$ & $\overline{S_{\rm s}}\uparrow$ & \rev{\makecell{Train Mem.~$\downarrow$}} & \rev{\makecell{Test Mem.~$\downarrow$}} \\ 
\midrule
    \textbf{L0} & 0 & 31.49 & 20.30 & 58.33 & 91.71 & \rev{21,160$^{*}$} & \rev{6,771$^{*}$} \\
    \textbf{L1} & 1 & 33.12$^{*}$ & 20.67$^{*}$ & 60.64 & 92.54 & \rev{37,804} & \rev{7,075} \\
    \textbf{L2} & 2 & 31.89 & 20.32 & 61.00$^{*}$ & 92.71$^{*}$ & \rev{53,298} & \rev{7,299} \\
    \textbf{L3} & 3 & 31.67 & 20.01 & 59.01 & 91.77 & \rev{68,770} & \rev{9,425} \\
    \textbf{L4} & 4 & \multicolumn{6}{c}{CUDA Out of Memory} \\ 
\bottomrule
\end{tabular}
\end{table}

When introducing historical keyframes (from $L=0$ to $L=1$), both occupancy accuracy metrics (IoU and mIoU) and temporal consistency metrics ($\overline{S_{\rm m}}$ and $\overline{S_{\rm s}}$) are improved, indicating that \ours enables SurroundOcc to leverage additional historical information and thus enrich temporal context. However, when $L$ is further increased to 2, although temporal consistency continues to improve, larger values of $L$ also introduce more irrelevant or noisy features. This makes it more challenging to associate historical information with the current occupancy prediction and may potentially lead to semantic misalignment.

Moreover, since \ours is designed for efficiency, increasing $L$ linearly expands the input volume and computational cost, leading to greater resource consumption during training and inference. \rev{Training memory grows almost exactly linearly, at about $15.1$~GB per additional keyframe; the measured value at $L=3$ differs from the linear fit through $L=1$ and $L=2$ by $0.03\%$, which confirms that the model is linear over this range. Continuing this trend to $L=4$ takes the requirement beyond the memory available on each of the L20 cards used in our experiments. At inference the growth is far smaller, between $0.2$ and $2.1$~GB per additional keyframe, so the window length constrains training considerably more than deployment.} Based on these findings, we recommend setting $L$ to 1 or 2 as a balanced choice between performance and efficiency.

\begin{table}[pos=!ht]
\centering
\caption{Effect of different aggregation strategies in ViewFormer+\ours. \rev{The memory was measured in MB on NVIDIA L20 GPUs.} The best result in each column is marked by an asterisk.}
\label{tab:ablation_fuse}
\begin{tabular}{ CCCCC }
\toprule
    Idx. & Strategy & mIoU$~\uparrow$ & Test Mem. $~\downarrow$ & Latency$~\downarrow$ \\
\midrule
    \textbf{S1} & Early aggregation  & 40.58 & 4,780 & 159 ms \\
    \textbf{S2} & Late aggregation & 41.30$^{*}$ & 4,687$^{*}$ & 121 ms$^{*}$ \\
\bottomrule
\end{tabular}
\end{table}

\noindent\textit{Impact of different aggregation strategies in \ours.}
As illustrated in \figurename~\ref{fig:cor}, \ours adopts a late aggregation strategy (\textbf{S2} in \tablename~\ref{tab:ablation_fuse}), where the learned tokens after MHAM are first passed through separate decoders and then aggregated in the occupancy space. In contrast, strategy \textbf{S1} employs early aggregation by combining the learned tokens in token space before decoding.

The results in \tablename~\ref{tab:ablation_fuse} show that the 3D-space fusion used in \textbf{S2} yields superior performance. This is because the output of \ours functions more effectively as an occupancy correction term when added directly in the occupancy space. Furthermore, the additive nature of \textbf{S2}'s design aligns more naturally with the plug-and-play paradigm than the early fusion approach used in \textbf{S1}.

\noindent\rev{\textit{Generality across backbones.}}
\rev{To verify that \ours does not overfit to a specific backbone, we replace the R101-DCN backbone of SurroundOcc with ViT-S~\cite{li2022exploring} and Swin-T~\cite{liu2024grounding}, keeping every other hyper-parameter and the training recipe unchanged. As \tablename~\ref{tab:backbone} shows, the gain is positive on all three backbones. \ours associates historical and motion cues with whatever features the backbone produces, and therefore improves VONs built on different backbone families consistently.}


\begin{table}[pos=!ht]
\centering
\caption{\rev{Effect of the image backbone. Performance gains ($\Delta$) are in \textbf{bold}.}}
\label{tab:backbone}
\begin{tabular}{ RCCCCCC }
\toprule
    \rev{\multirow{2}{*}[-0.4em]{Method}} & \multicolumn{2}{c}{\rev{R101-DCN}} & \multicolumn{2}{c}{\rev{ViT~\cite{li2022exploring}}} & \multicolumn{2}{c}{\rev{Swin-T~\cite{liu2024grounding}}} \\
\cmidrule(lr){2-3} \cmidrule(lr){4-5} \cmidrule(lr){6-7}
     & \rev{IoU~$\uparrow$} & \rev{mIoU~$\uparrow$} & \rev{IoU~$\uparrow$} & \rev{mIoU~$\uparrow$} & \rev{IoU~$\uparrow$} & \rev{mIoU~$\uparrow$} \\
\midrule
    \rev{SurroundOcc} & \rev{31.49} & \rev{20.30} & \rev{27.97} & \rev{12.35} & \rev{31.05} & \rev{17.93} \\
    \rev{\quad + \ours} & \rev{33.12} & \rev{20.67} & \rev{28.22} & \rev{12.51} & \rev{31.38} & \rev{18.16} \\
    \rev{$\Delta$} & \rev{\textbf{+1.63}} & \rev{\textbf{+0.37}} & \rev{\textbf{+0.25}} & \rev{\textbf{+0.16}} & \rev{\textbf{+0.33}} & \rev{\textbf{+0.23}} \\
\bottomrule
\end{tabular}
\end{table}

\noindent\rev{\textit{Comparison with training-free temporal filtering.}}
\rev{\tablename~\ref{tab:postproc} compares \ours with a training-free steady-state Kalman filter applied to the same SurroundOcc predictions, equivalent to a per-voxel exponential moving average with observation gain $\alpha$, with and without ego-motion compensation. The filter trades accuracy for consistency: driving $\alpha$ down to $0.15$ pushes $\overline{S_{\rm m}}$ to $89.16$ while mIoU collapses to $11.84$, so the consistency scores must be read jointly with accuracy. The filter processes each voxel independently, carries no learned prior, and can only redistribute what the base network already predicts; ego-motion compensation recovers geometric overlap but supplies no semantic evidence. \ours instead learns structural priors of objects and the spatial continuity of the road surface, draws new evidence from historical keyframes and motion cues, and is the only entry in the table that improves accuracy and consistency together.}

\begin{table}[pos=!ht]
\centering
\caption{\rev{Training-free temporal filtering applied to the SurroundOcc predictions, compared with \ours. The standard and ego-motion-compensated filters use $\alpha=0.45$ and $\alpha=0.35$, respectively, the most favourable setting for each variant. Values are recomputed from dumped predictions and differ from \tablename~\ref{tab:main-res-a} by at most $0.7$. The best result in each column is marked by an asterisk.}}
\label{tab:postproc}
\begin{tabular}{ RCCCC }
\toprule
    \rev{Method} & \rev{IoU~$\uparrow$} & \rev{mIoU~$\uparrow$} & \rev{$\overline{S_{\rm m}}\uparrow$} & \rev{$\overline{S_{\rm s}}\uparrow$} \\
\midrule
    \rev{SurroundOcc} & \rev{31.45} & \rev{20.26} & \rev{58.06} & \rev{92.37} \\
    \rev{\quad + temporal filtering ($\alpha=0.45$)} & \rev{30.50} & \rev{17.67} & \rev{67.22$^{*}$} & \rev{94.69$^{*}$} \\
    \rev{\quad + ego-motion-comp.\ filtering ($\alpha=0.35$)} & \rev{32.09} & \rev{20.13} & \rev{61.66} & \rev{93.72} \\
    \rev{\quad + \ours} & \rev{33.12$^{*}$} & \rev{20.67$^{*}$} & \rev{60.64} & \rev{92.54} \\
\bottomrule
\end{tabular}
\end{table}

\begin{figure}[pos=!htbp]
\centering
\includegraphics[width=0.85\textwidth]{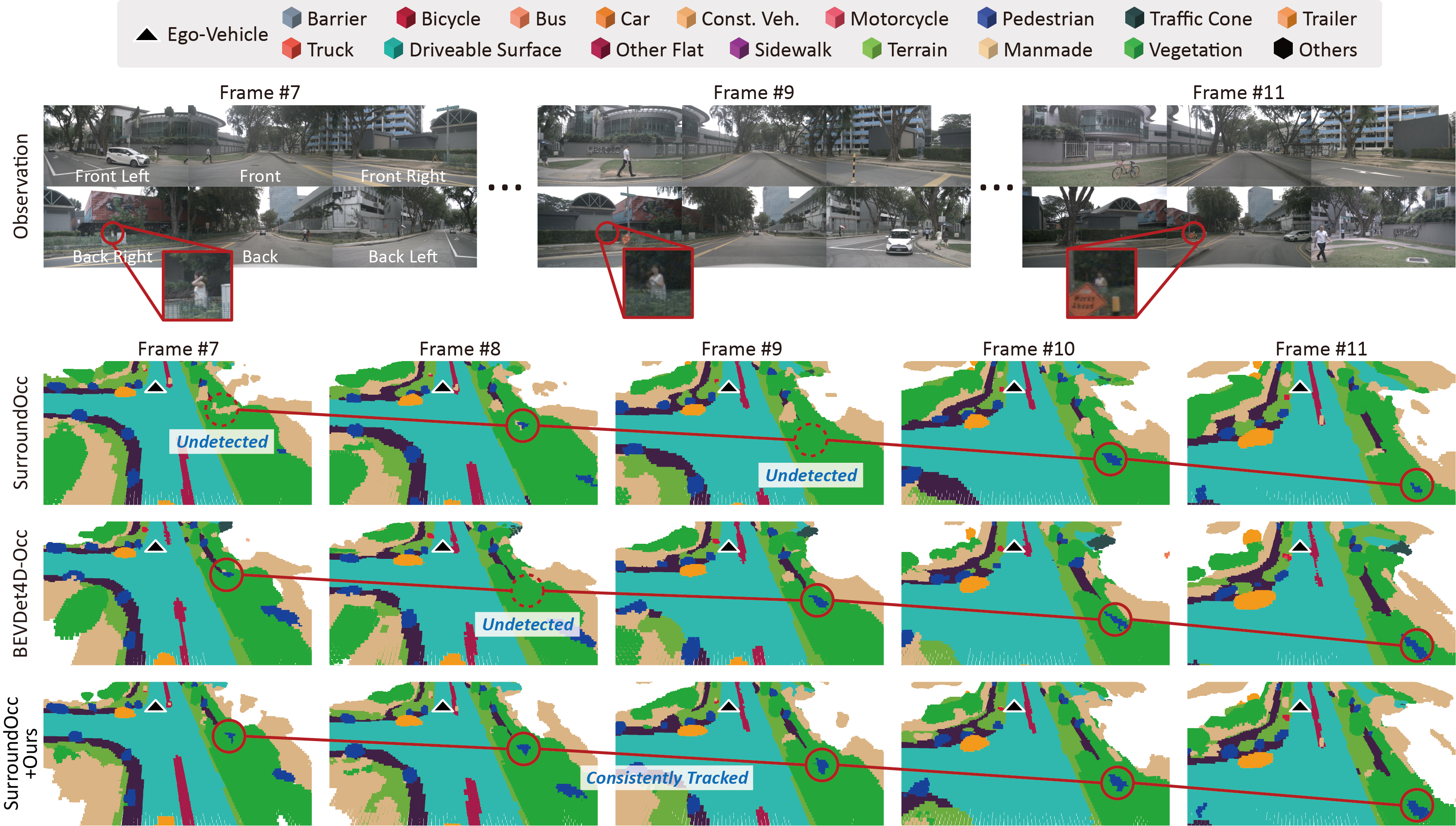}
\caption{Comparison in a T-junction scenario with partial and dynamic pedestrian occlusions. SurroundOcc+\ours produces robust predictions, with the pedestrian being consistently tracked, while SOTA methods show a flickering phenomenon.}
\label{fig:case-study-big}
\end{figure}

\begin{figure}[pos=!htbp]
\centering
\includegraphics[width=0.7\linewidth]{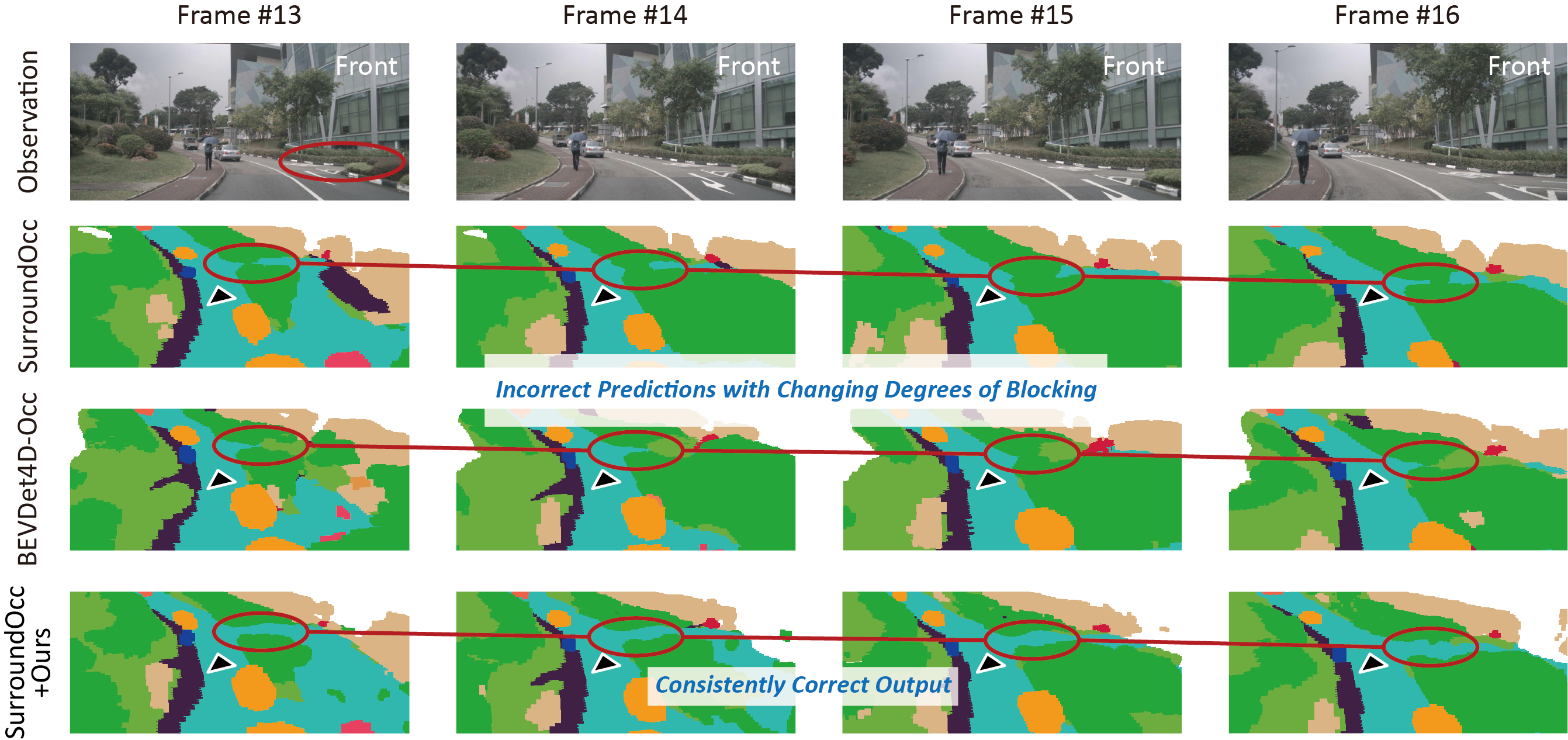}
\caption{Comparison in a ramp entry scenario with partial occlusions from roadside vegetation. SurroundOcc+\ours produces robust and continuous predictions of the drivable surface, while SOTA methods occasionally misclassify the ramp as blocked.}
\label{fig:case-study-big2}
\end{figure}

\subsection{Case analysis}
To qualitatively assess the effectiveness of our method, we compare SurroundOcc+\ours with a single-frame 3D VON baseline (SurroundOcc~\cite{surroundOcc}) and a history-aware VON method (BEVDet4D-Occ~\cite{bevdet4d}).

\noindent\textit{Temporal visualization cases.}
As shown in~\figurename~\ref{fig:case-study-big} (Scene 277, Frames \#7–\#11), a pedestrian walking along the sidewalk -- parallel to the ego vehicle's trajectory -- is intermittently occluded by roadside vegetation. SurroundOcc~\cite{surroundOcc} exhibits significant instability, with the pedestrian missing in Frames \#7 and \#9, highlighting its limitations in temporal modeling. BEVDet4D-Occ~\cite{bevdet4d} partially alleviates this issue by fusing historical latent features into the current latent space, but still shows occasional inconsistencies, such as a detection dropout in Frame \#8. In contrast, SurroundOcc+\ours completely eliminates these flickering artifacts and achieves consistent detection across all occlusion states.

\begin{figure}[pos=!htbp]
\centering
\includegraphics[width=0.85\textwidth]{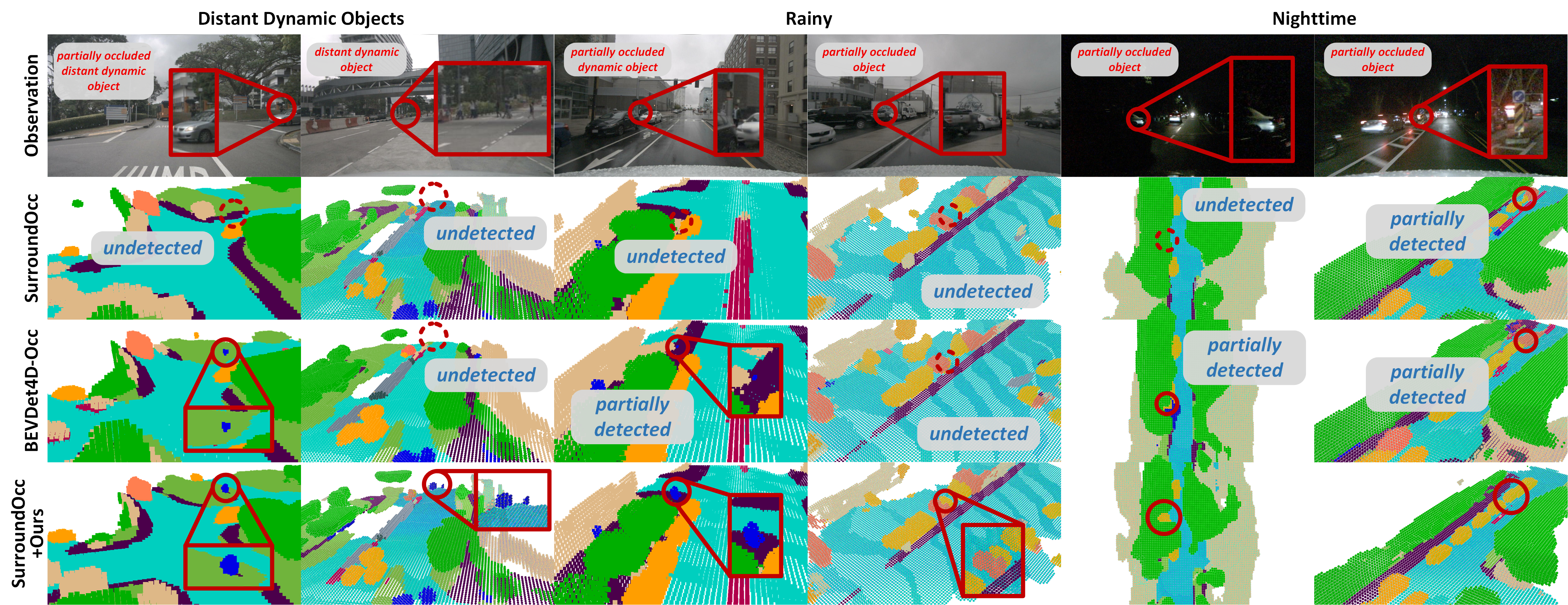}
\caption{\rev{Visualization of three challenging conditions, each containing two cases. A dashed circle marks a target whose class is absent from the prediction, while a solid circle marks one that is present. SurroundOcc+\ours recovers targets that the baselines miss or only partially recover.}}

\label{fig:case-extra}
\end{figure}

As shown in \figurename~\ref{fig:case-study-big2} (Scene 926, Frames \#13-\#16), in the ramp entry scenario, the front-right ramp lane in the camera front view is partially occluded by roadside vegetation. SurroundOcc+\ours produces robust and continuous predictions of the drivable surface at the junction, while the baseline methods exhibit flickering predictions, occasionally misclassifying the entrance as blocked.

\noindent\textit{Visualization of challenging scenarios.}
\rev{\figurename~\ref{fig:case-extra} covers three challenging conditions, with two cases included for each condition. (1) Distant dynamic objects may be occluded, resulting in sparse visual cues. (2) In rainy scenes, the images may become wet and blurry due to the weather, and water may accumulate on the road. (3) Under extreme nighttime lighting conditions, the images may be underexposed, resulting in strong contrast between bright and dark regions. Therefore, partially occluded objects are more likely to be missed by traditional occupancy prediction methods, whereas our method can effectively enhance occupancy prediction and maintain its accuracy.}

\begin{table}[pos=!ht]
\centering
\caption{Comparison of computational overhead. All models share the same visual encoder ($\Phi_{\rm SE}$) of ResNet-50 backbones. Our result (ViewFormer+\ours) in this table is measured for $L=1$. OOM indicates CUDA out of memory. \rev{Latency is measured on both an NVIDIA L20 and an NVIDIA A100.} Performance changes ($\Delta$) are highlighted in \textbf{bold}, and best result in each column is marked by an asterisk.}
\label{tab:efficiency}
\begin{tabular}{ RCCCC }
\toprule
    \multirow{2}{*}[-0.2em]{Model} & \multirow{2}{*}[-0.2em]{mIoU~$\uparrow$} & \multirow{2}{*}[-0.2em]{\makecell{Memory\\(MB)~$\downarrow$}} & \multicolumn{2}{c}{Latency~$\downarrow$} \\
\cmidrule(lr){4-5}
     & & & L20 & \rev{A100} \\
\midrule
    FB-Occ~\cite{fb_occ} & 39.11 & \makecell[r]{5,933} & 0.09 s & \rev{0.13 s} \\
    SparseOcc~\cite{SparseOcc_Liu} & 30.10 & \makecell[r]{7,147} & 0.05 s & \rev{0.05 s} \\
    OPUS-L~\cite{opus} & 36.20 & \makecell[r]{10,579} & 0.16 s & \rev{0.09 s} \\
    OPUS-T~\cite{opus} & 33.20 & \makecell[r]{6,711} & 0.03 s$^{*}$ & \rev{0.04 s$^{*}$} \\
    BEVDet4D-Occ~\cite{bevdet4d} & 39.30 & \makecell[r]{4,689} & 0.26 s & \rev{0.29 s} \\
\midrule
    ViewFormer~\cite{viewformer} & 40.46 & \makecell[r]{4,582$^{*}$} & 0.11 s & \rev{0.13 s} \\ 
    \makecell[r]{+\ours\\$\Delta$} & \makecell[c]{41.30$^{*}$\\\textbf{+0.84}} & \makecell[c]{4,687\\\textbf{+105}} & \makecell[c]{0.12 s\\\textbf{+0.01 s}} & \rev{\makecell[c]{0.14 s\\\textbf{+0.01 s}}} \\
\bottomrule
\end{tabular}
\end{table}

\subsection{Efficiency–performance tradeoff analysis}

To ensure a fair comparison, all overhead analysis experiments are conducted on a single NVIDIA L20 GPU. \tablename~\ref{tab:efficiency} summarizes the results of several state-of-the-art methods, including history-aware approaches (FB-Occ~\cite{fb_occ}, SparseOcc~\cite{SparseOcc_Liu}, OPUS-L/M/S/T~\cite{opus}, and BEVDet4D-Occ~\cite{bevdet4d}), the single-frame 3D VON baseline (ViewFormer~\cite{viewformer}), and our method combined with ViewFormer. All models are evaluated on the Occ3D benchmark~\cite{Occ3D} using the same 2D visual encoder backbone, ResNet-50, which outputs feature maps with a resolution of $(32, 88)$. The table reports the mIoU, GPU memory usage during inference, and per-sample latency of each method.

The results show that ViewFormer+\ours achieves state-of-the-art mIoU while maintaining minimal GPU memory usage. The \ours plug-in consumes only 105MB, accounting for just 2.29\% of ViewFormer's memory footprint, and introduces a mere 0.01 seconds increase in inference latency. These findings confirm that \ours delivers competitive performance with negligible computational overhead. 

\section{Conclusion}
\label{sec:con}
This paper introduces \ours, a lightweight plug-and-play module that reduces temporal flickering in VONs without modifying or retraining the base network. It encodes historical static features and frame-difference motion cues as sparse tokens, models their spatio-temporal correlations through dual cross-attention, and adds the learned correction to the base occupancy logits. We also introduced $\overline{S_{\rm m}}$ and $\overline{S_{\rm s}}$ based on MOC/SOC voxel-change categories to evaluate consistency in moving and static regions. Across the nuScenes-derived SurroundOcc~\cite{surroundOcc} and Occ3D~\cite{Occ3D} benchmarks, \ours improves SurroundOcc's IoU by 1.63\% and $\overline{S_{\rm m}}$ by 2.31, and ViewFormer's mIoU by 0.84\%, while adding only 105MB of GPU memory and 0.01s of latency per frame. \rev{Temporal cues are aligned implicitly in the latent space without explicit ego-pose transformation}, making \ours applicable to both single-frame and history-aware VONs.

\noindent\textit{Strengths and limitations.}
\rev{The key strengths of \ours are its modularity, efficiency, and interpretable static-motion decomposition. Nevertheless, temporal context is limited by GPU memory ($L=4$ causes OOM in \tablename~\ref{tab:ntrack}), while longer windows may introduce semantic misalignment. Performance also depends on the frozen base features, frame differences may be unreliable under large ego-motion, and generalization beyond nuScenes-derived benchmarks remains untested.}

\noindent\textit{Future work.}
\rev{We will explore memory-efficient attention for longer temporal context, learned fusion of frame differences and optical flow, and validation on additional datasets. Jointly fine-tuning the base VON with \ours may further improve accuracy, although it would relax the plug-and-play property. \rev{Drastic illumination changes, such as tunnel transitions, also remain challenging because frame differencing is sensitive to global brightness shifts. We will explore DVS-based sensing~\cite{xu2026decomposed} and collect occupancy data under extreme illumination conditions.} We hope \ours provides a practical basis for efficient, history-aware occupancy deflickering.}

\bibliographystyle{cas-model2-names}

\renewcommand{\bibfont}{\small}

\bibliography{cas-refs}

\end{document}